\pdfoutput=1
\documentclass[11pt]{article}

\usepackage[final]{acl}

\usepackage{times}
\usepackage{latexsym}

\usepackage[T1]{fontenc}
\usepackage[utf8]{inputenc}

\usepackage{microtype}

\usepackage{inconsolata}

\usepackage{tabularx}
\usepackage{caption}
\usepackage{array}
\usepackage{xcolor}

\usepackage{booktabs}
\usepackage{times}
\usepackage{helvet}
\usepackage{courier}
\usepackage{graphicx}
\usepackage{float}
\usepackage[switch]{lineno}
\usepackage{multirow}
\usepackage{adjustbox}
\usepackage{amsmath}
\usepackage{amssymb}
\usepackage{xcolor}
\usepackage{algorithmic}
\usepackage[ruled,vlined]{algorithm2e}
\usepackage{subcaption}
\usepackage{kotex} 
\usepackage{comment}

\usepackage{amsmath}
\usepackage{amssymb}
\usepackage{mathtools}
\usepackage{amsthm}
\usepackage{bbm}

\usepackage{kotex}
\usepackage{multirow}
\usepackage{marvosym}
\usepackage{makecell}
\usepackage{comment}
\usepackage{enumitem}
\usepackage{enumerate}
\usepackage{makecell}
\usepackage{pifont}

\DeclareMathOperator*{\argmax}{argmax}
\newcommand{\oursol}{\textsc{CoREn}}
\newcommand{\lmgen}{\Phi_{\textnormal{LLM}}}

\usepackage{graphicx}

\title{LLM-Based Offline Learning for Embodied Agents via  Consistency-Guided Reward Ensemble}

\author{Yujeong Lee\textsuperscript{\rm 1}\thanks{Equally contributed to this work},
Sangwoo Shin\textsuperscript{\rm 1}\footnotemark[1], 
Wei-Jin Park\textsuperscript{\rm 2}, 
Honguk Woo\textsuperscript{\rm 1}\thanks{Corresponding author
}
\\
    Department of Computer Science and Engineering, Sungkyunkwan University\textsuperscript{1} \\
    Acryl Inc.\textsuperscript{2} \\
    \texttt{yujlee@skku.edu, jsw7460@skku.edu, jin@acryl.ai, hwoo@skku.edu}
  }

\begin{document}
\maketitle
\begin{abstract}
Employing large language models (LLMs) to enable embodied agents has become popular, yet it presents several limitations in practice. In this work, rather than using LLMs directly as agents, we explore their use as tools for embodied agent learning. Specifically, to train separate agents via offline reinforcement learning (RL), an LLM is used to provide dense reward feedback on individual actions in training datasets. 
In doing so, we present a consistency-guided reward ensemble framework ($\oursol$), designed for tackling difficulties in grounding LLM-generated estimates to the target environment domain.
The framework employs an adaptive ensemble of spatio-temporally consistent rewards to derive domain-grounded rewards in the training datasets, thus enabling effective offline learning of embodied agents in different environment domains. 
Experiments with the VirtualHome benchmark demonstrate that $\oursol$ significantly outperforms other offline RL agents, and it also achieves comparable performance to state-of-the-art LLM-based agents with 8B parameters, despite $\oursol$ having only 117M parameters for the agent policy network and using LLMs only for training.
\end{abstract}

\section{Introduction}
Developing embodied agents capable of understanding user instructions and executing tasks in physical environments represents a crucial milestone in the pursuit of general AI. Recent advancements in large language models (LLMs) have demonstrated their remarkable reasoning capabilities, paving the way for their application in embodied agents~\cite{LACMA, padmakumar2023multimodal, EMMA, yun2023emergence, logeswaran2022few, saycan}. 
Yet, deploying an LLM directly as part of an embodied agent presents several inefficiencies, such as the need for sophisticated environment-specific prompt design, substantial computational resource demands, and inherent model inference latency~\cite{subgoal_distill}. These factors can limit the practical application of LLMs, particularly in scenarios where embodied agents are required to respond rapidly and efficiently.

In the literature of reinforcement learning (RL), data-centric offline learning approaches have been explored~\cite{levine_offrl}. These offline RL approaches are designed to establish efficient agent structures, necessitating datasets that include well-annotated agent trajectories with reward information. However, the characteristics of instruction-following tasks assigned to embodied agents, particularly their long-horizon goal-reaching nature, often conflict with such dense data requirements of offline RL. Embodied agents normally can produce trajectories with sparse reward feedback, because their instruction-following tasks are evaluated based on binary outcomes of success or failure, which directly align with the specific goals of the instructions. In offline RL, this sparse reward setting poses significant challenges in achieving effective agent policies~\cite{park_hiql, jason_offgoal_f}.

In this work, we explore LLMs for offline RL.
By employing capable LLMs as a reward estimator that provides immediate feedback on agent actions, we augment the trajectory dataset with dense reward information. This method, \textbf{LLM-based reward estimation} is capable of significantly enhancing the effectiveness of offline RL for embodied agents. 
To do so, we address the limitations inherent in LLM-based reward estimation. 
A primary challenge arises from the limited interaction with the environment in an offline setting, which complicates the LLMs' ability to acquire essential environmental knowledge. The offline setting makes it difficult to ensure that the generated rewards are properly grounded in the specific domain of the environment.
\begin{figure}[t]
    \centering    
    \includegraphics[width=0.48\textwidth]{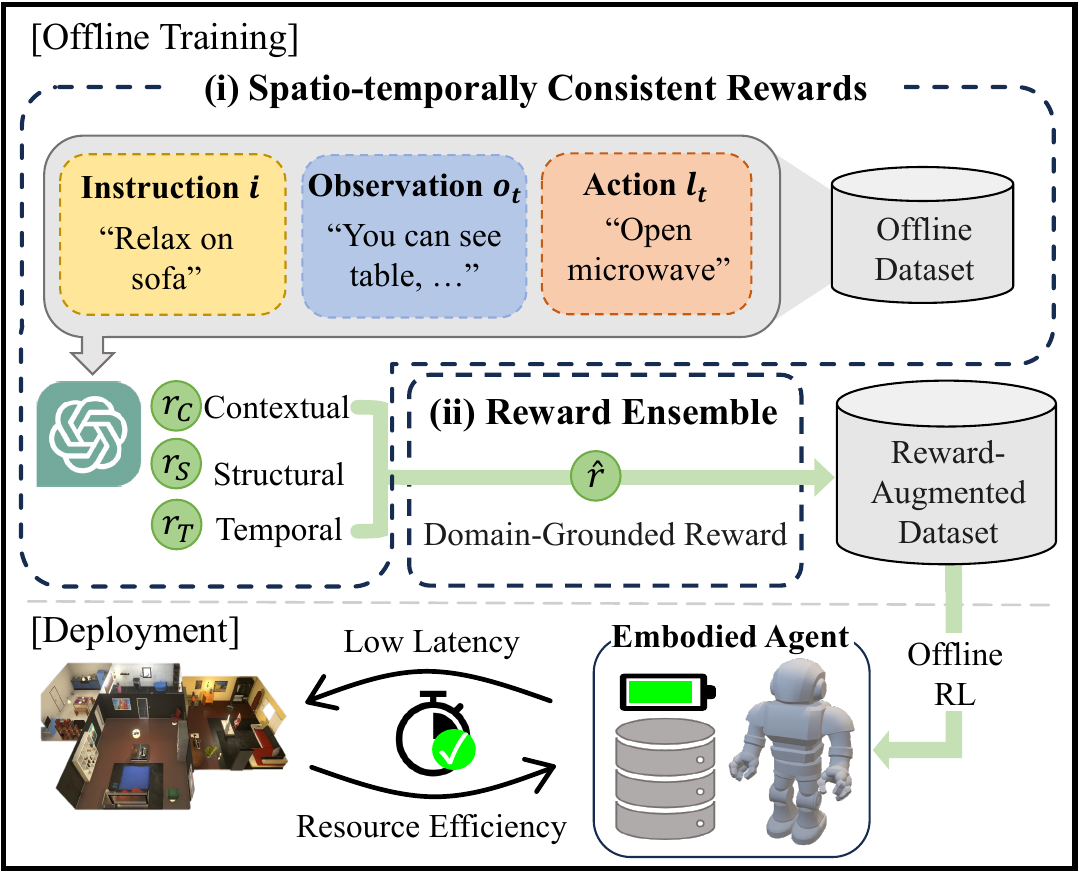}
    \caption{$\oursol$, a framework for LLM-based reward estimation and offline learning.  
    In (i), an LLM estimates rewards  based on spatio-temporal (i.e., contextual, structural, and temporal) consistencies; 
    In (ii), these rewards are integrated into a single domain-grounded reward via an ensemble. Using the reward-augmented dataset, offline RL can be conducted effectively to achieve embodied agents with resource efficiency and low latency.
    }
    \label{fig:fig1}
\end{figure}
% 
% Camera Ready: 예시만 변경함 (이게 좀 더 그럴듯 해보임)
For instance, without explicit knowledge that a flowerpot is typically stored in a living room in the target environment, an LLM might struggle to accurately assign rewards for actions like ``go to living room'' versus ``go to balcony'' when tasked with watering plants. While both actions might seem reasonable from a commonsense perspective, the optimal action depends on specific conditions of the target environment that the LLM may not have access to in the offline setting.

% 0921: 여기 citation 추가 (리뷰어 xyuZ에게 한 첫번째 답변 참고)
These challenges, unique to the \textit{offline} context, differentiate our work from previous works on \textit{online} LLM-based reward estimation, where LLMs can be fine-tuned or prompts can be refined through repeated interaction with environment or human~\cite{InstructPatentGPT, Text2Reward, Auto_MC_Reward, Automated_Reward_Function_Designer}. Since these interactions are not available in offline settings, improving the LLM's insufficient spatial reasoning for accurate reward estimation requires a fundamentally different approach.
\color{black}

In response, we present \textbf{$\oursol$}, a consistency-guided reward ensemble framework, specifically designed for robust LLM-based reward estimation and effective agent offline learning. 
%%%%%%%%%%%%%%%%%%%%%%%%%%%%%%%%%
It adopts a two-staged reward estimation process, as depicted in Figure~\ref{fig:fig1}.
%
% consistency 이름을 명시하면서 좀 더 자세히 써야하나...
(i) An LLM is first queried to estimate several types of rewards for actions, each considering a distinct spatio-temporal consistency criterion of the LLM to have coherent and domain-grounded rewards.
(ii) Then, these rewards are further orchestrated, being unified into domain-specifically tuned rewards via an alignment process with the sparse rewards of given trajectories. 
The resulting agent, trained on the unified dense rewards by offline RL, is capable of performing instruction-following tasks with high efficiency and minimal latency at deployment. This offline RL scheme, enhanced by LLM-based reward estimation,  overcomes the limitations faced by the agents that rely on the online exploitation of LLMs.
%%%%%%%%%%%%%%%%%%%%%%%%%%%%%%%

%%
The contributions of our work can be summarized as follows: 
\textbf{(i)} addressing a practical yet challenging problem of embodied agent \textit{offline} learning using LLMs for the first time; 
\textbf{(ii)} proposing a two-staged reward estimation algorithm guided by a spatio-temporal consistency ensemble; and 
\textbf{(iii)} extensive evaluation on the VirtualHome benchmark, demonstrating performance comparable to state-of-the-art LLM-based online agents.

\section{Preliminaries}
\subsection{Goal-POMDPs}
For an embodied agent that follows user-specified instructions, we model their environment as a goal-conditioned partially observable Markov decision process (Goal-POMDP). 
% The optimal policy maps combined observations to actions, i.e., $\pi^*: \hat{o} = (i, o) \mapsto a \in \mathcal{A}$. 
%
A Goal-POMDP is represented by a tuple ($\mathcal{S}$, $\mathcal{A}$, $P$, $R$, $\gamma$, $\Omega$, $\mathcal{O}$, $\mathcal{G}$)~\cite{llm-planner, progprompt} 
with 
states $s \in \mathcal{S}$, 
actions $a \in \mathcal{A}$, 
a transition function $P: \mathcal{S} \times \mathcal{A}  \longrightarrow \Delta(\mathcal{S})$, 
a reward function $R : \mathcal{S} \times \mathcal{A} \times \mathcal{G}^\omega\footnote{$X^\omega$ for a set $X$ is all possible finite products of $X$.} \mapsto \mathbb{R}$, 
a discount factor $\gamma \in [0, 1)$, 
observations $o \in \Omega$, 
an observation transition function $\mathcal{O}: \mathcal{S} \times \mathcal{A} \longrightarrow \Omega$, and 
goal conditions $G \in \mathcal{G}$. 
Given this Goal-POMDP representation, we consider a user-specified instruction $i$ as a series of goal conditions $\mathbf{G} = (G_1, \cdots ) \subseteq \mathcal{G}$ such that the embodied agent is tasked with completing each of the specified goal conditions for the instruction $i$. 

\begin{figure*}[t]
    \centering
     \includegraphics[width=1.0\textwidth]{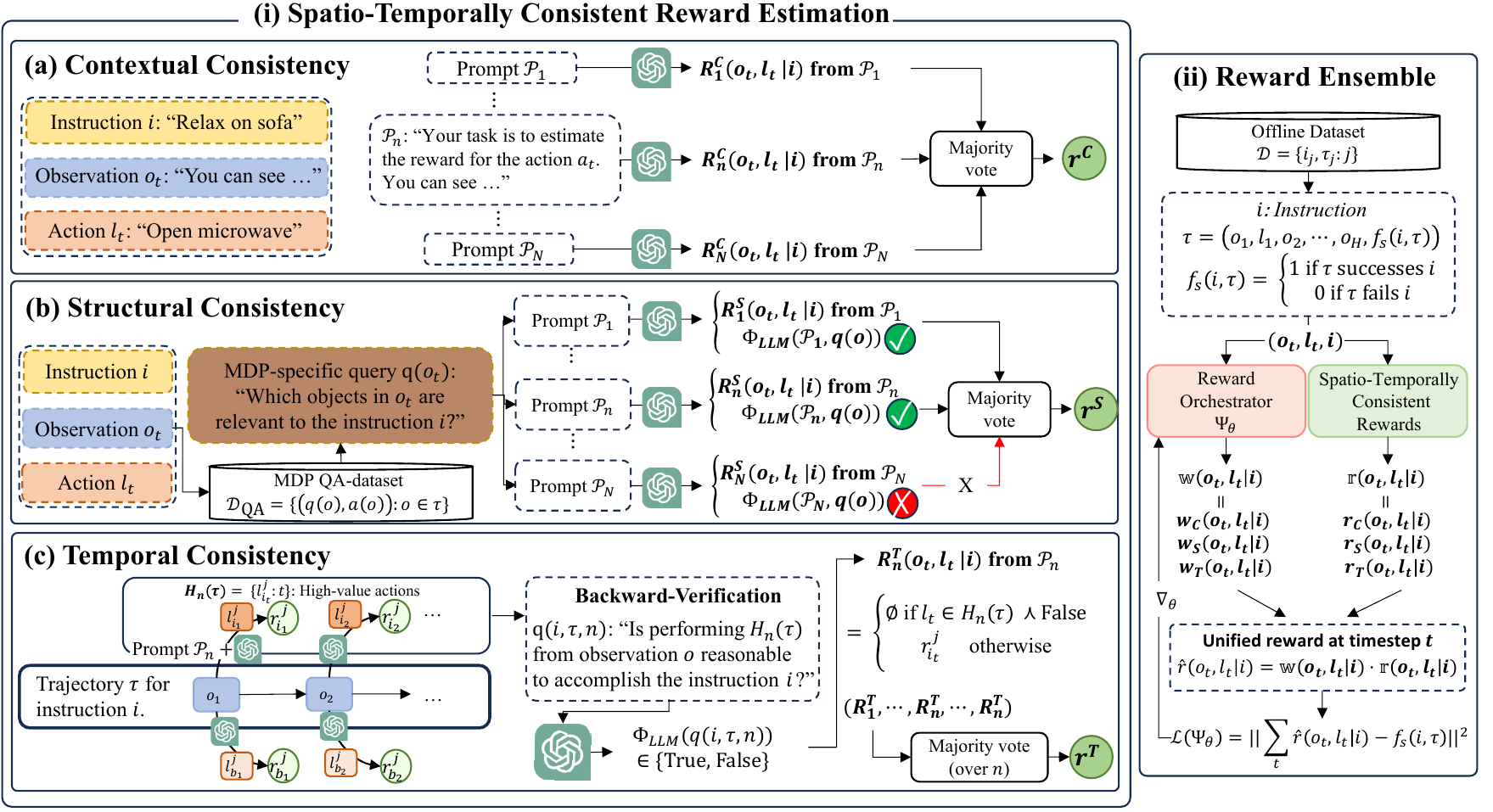}
    \caption{
    Two-staged reward estimation in $\oursol$. 
    In (i), spatio-temporally consistent rewards, constrained by contextual, structural, and temporal consistencies, are calculated.
    (a) Contextual consistency is achieved through majority voting across the responses from different prompts $\mathcal{P}_n$, resulting in contextually consistent rewards $r^C$.
    (b) Structural consistency is achieved by presenting MDP-specific queries to the LLM. If the LLM incorrectly answers these queries (indicated by a red `X'), the rewards estimated from these particular prompts are removed from majority voting. The successfully verified rewards contribute to structurally consistent rewards $r^S$.
    (c) Temporal consistency involves collecting high-value actions $H_n(\tau)$ and subjecting them to backward verification through LLM queries. Actions that fail this verification are excluded from the candidates for majority voting. Otherwise, they contribute to temporally consistent rewards $r^T$.
    In (ii), a trajectory $(i, \tau)$ with success flag $f_s(i, \tau)$ is sampled from the given offline dataset $\mathcal{D}$. The spatio-temporally consistent rewards $(r^C, r^S, r^T)$ in (i) are combined using weights $(w^C, w^S, w^T)$, which are generated by the reward orchestrator $\Psi_\theta$.
    This combined result renders a unified stepwise, more domain-grounded reward $\hat{r}$. The orchestrator $\Psi_\theta$ is trained to align the trajectory's return of accumulating stepwise rewards $\hat{r}$ with the sparse reward $f_s(i, \tau)$ annotated on the trajectory.
    }
    \label{fig:fig2}
\end{figure*}
\subsection{Offline RL}
For a Goal-POMDP, its optimal policy is formulated by
\begin{equation}
\label{eq: optimal policy}
    \pi^* = \argmax_{\pi} \underset{\substack{(s, a) \sim \pi, \\ \mathbf{G} \sim \mathcal{G}}}{\mathbb{E}} \left[\sum_t \gamma^t R(s, a, \mathbf{G})\right].
\end{equation}
To achieve the optimal policy, we explore offline RL approaches where the policy is derived by optimizing the Bellman error objective, relying exclusively on an offline dataset $\mathcal{D}$ without any environment interaction.
Offline RL is particularly beneficial for embodied agents, as it reduces the risks and costs associated with active exploration of the environment with physical objects.
We utilize $\mathcal{D} = \{(i_j, \tau_j): j \}$ where $\tau_j$ is a trajectory corresponding to instruction $i_j$. Unlike conventional offline RL, this dataset $\mathcal{D}$ incorporates sparse rewards. This sparsity is reflected in a subset of trajectories that are marked by a success flag $f_s(i_j, \tau_j)$, indicating whether $\tau_j$ has satisfied all the requisite goal conditions for the instruction $i_j$. This sparse reward setup is inherent for embodied instruction-following tasks, as each instruction is treated as a series of goal conditions within Goal-POMDPs. 
%
% However, this sparse reward setup poses significant challenges in adopting offline RL approaches~\cite{jason_offgoal_f, park_hiql}. Learning a robust policy in an environment with expansive state-goal spaces and restricted interaction is inherently difficult, and the difficulty is intensified by the sparsity of reward feedback.

% \subsection{LLM-based Reward Estimation}

\section{Our Approach}
\noindent\textbf{LLM-based reward estimation.} 
Offline RL facilitates agent learning without direct environment interaction, but relying solely on sparse rewards to learn long-horizon instruction-following tasks is often inefficient. 
To improve this, we augment agent trajectories with stepwise intrinsic rewards through LLM-based estimation. 
Similar to LLM-based task planning~\cite{progprompt, saycan}, LLMs can be used to approximate the reward of observation-action pairs in the dataset, providing more immediate and actionable dense feedback to enhance the effectiveness of offline learning.
%
%To do so, we address the following limitations of LLM-based reward estimation.  

\noindent\textbf{Not-grounded reward estimation.} 
Intrinsic rewards estimated by LLMs at intermediate steps might not consistently align with the sparse rewards provided at the conclusion of individual instruction-following tasks. This discrepancy arises when the intrinsic rewards are not sufficiently grounded in the environment domain. 
This issue is exacerbated in a partially observable setting, where LLMs are forced to infer rewards based on incomplete snapshots of the environment.

\subsection{Overall Framework}
To tackle the limitations of LLM-based reward estimation, we propose a spatio-temporal consistency-guided reward ensemble framework $\oursol$ with a two-stage process.  
As described in Figure~\ref{fig:fig2}, the first stage (i) incorporates contextual, structural, and temporal consistencies to fully harness the LLM's reasoning ability and enhance the groundedness of reward estimates within the specific domain of the embodied environment. 
In the second stage (ii), $\oursol$ orchestrates an ensemble of distinct rewards generated during the first stage based on the trajectories' success. This allows for the derivation of domain-specifically tuned rewards, which can be effectively utilized for the offline learning of embodied agents.

\subsection{Spatio-Temporally Consistent Rewards}
\label{sec: consistent rewards}
For reward estimation, we employ $N$ distinct prompts $\mathcal{P}_1, \cdots, \mathcal{P}_N$ with an LLM ($\lmgen$), where a prompt is distinguished by its unique explanations, in-context demonstrations, as well as the use of a chain-of-thought (CoT). 
Specifically, each prompt $\mathcal{P}_n$ combined with observation $o$, action $l$, and instruction $i$ is used to generate rewards $R_n$ through $\lmgen$ inferences.
\begin{equation}
    R_{n}(o, l | i) = \lmgen(P_n, (o, l, i))
    \label{eq:reward}
\end{equation}

%\subsubsection{Spatial Consistency}
%
Spatial consistency is intended to ensure that the domain-grounded LLM’s reward estimation remains consistent across different prompt-induced contexts as well as it is based on a comprehensive understanding of the environmental structure. We achieve this using the implementation of two consistency mechanisms. 
%This is achieved based on contextual and structural consistencies.

\noindent\textbf{Contextual consistency.}
This mechanism aims to mitigate biases stemming from specific prompt contexts used in LLM-based reward estimation. By employing multiple $N$ prompts, each with a different contextual frame, we ensure that the rewards, which remain consistent across these variations, reflect a consensus in reasoning. 
For contextually consistent rewards $r_C$, we integrate the responses $R_n^C(o, l | i)$ of  prompts $\mathcal{P}_n$ by 
\begin{equation}
\label{eq: majority consistency}
    % \begin{cases}
        r^C(o, l | i) = \underset{r \in \hat{\mathbb{R}}}{\text{argmax}} \sum\limits_{n=1}^N \mathbbm{1}_{(R_n^C(o, l | i) = r)} \\
    % \end{cases}
\end{equation}
where $R_n^C(o, l | i) = \lmgen(\mathcal{P}_n, (o, l, i))$.

\noindent\textbf{Structural consistency.}
This is intended to ensure that the reward estimation incorporates a comprehensive understanding of the environment physical structure, such as objects, their relationships, and their relevance to the given instruction. 
We inquire $\lmgen$ with MDP-specific queries $q(o)$ relevant to observation $o$ such as ``Which objects in $o$ are relevant to the instruction $i$?''.
Exploiting the response $\lmgen(\mathcal{P}_n, q(o))$ to these queries, we integrate the rewards $R_n^S(o, l | i)$ of prompts $\mathcal{P}_n$:
\begin{equation}
\label{eq: structural consistency}
    r^S(o, l | i) = \underset{r \in \hat{\mathbb{R}}}{\text{argmax}} \sum_{n=1}^N \mathbbm{1}_{(R_n^S(o, l | i) = r)}.
\end{equation}
We rewrite Eq.~\eqref{eq:reward} for query violation cases, obtaining 
$R_n^S(o, l | i)$
\begin{equation}
\resizebox{1.0\hsize}{!}{
\label{eq: structural candidate}
$=
    \begin{cases}
        \varnothing & a(o) \neq \lmgen(\mathcal{P}_n, q(o)) \\
        \lmgen(\mathcal{P}_n, (o, l, i)) & \textnormal{otherwise}.
    \end{cases}
$
}
\end{equation}
%
% Here, for an MDP-specific query $q(o)$, $E$ evaluates whether the response $\lmgen(q(o))$ matches its corresponding ground-truth answer $a(o)$.
%
Details of prompts $\mathcal{P}_n$ and the dataset construction for MDP-specific queries and answers $\mathcal{D}_{\textnormal{QA}} = \{(q(o), a(o)): o \in \tau \in \mathcal{D} \}$ are in Appendix.

%\subsubsection{Temporal Consistency}
\noindent\textbf{Temporal consistency.}
This is designed to ensure that the value assigned to an action remains coherent throughout its whole decision-making process. 
With temporal consistency, if forward reasoning by the LLM assesses certain actions as having high values, backward verification must confirm that these high-value actions can collectively accomplish the given instruction.

To achieve this backward verification, we inquire $\lmgen$ with the query $q(i, \tau, n)$: ``Is performing the high-value actions $H_n(\tau)$ from observation $o$ reasonable to accomplish the instruction $i$?''. 
% --> y hat 같은 문자 전부 없애고, LLM 표기만 사용해서 input-output으로만 정답 매기는 것으로 했습니다.
The reward is then contingent on the response $\lmgen(q(i, \tau, n)) \in \{\textnormal{True}, \textnormal{False}\}$ to this query, and Eq.~\eqref{eq:reward} is rewritten as $R_n^T(o, l | i)$
\begin{equation}
\resizebox{1.0\hsize}{!}{
\label{eq: temporal candidate}
$
\begin{split}
=
\begin{cases}
\varnothing & l \in H_n(\tau) \wedge \neg \lmgen(q(i, \tau, n)) \\
\lmgen(\mathcal{P}_n, (o, l, i)) & \textnormal{otherwise}
\end{cases}
\end{split}
$
}
\end{equation}
for the cases of query violation, i.e., $l \in H_n(\tau) \wedge \neg \lmgen(q(i, \tau, n))$. 
Here, for all trajectory observations $o \in \tau$, high-value actions are defined as 
\begin{equation}
\label{eq: highvalue-actions}
    H_n(\tau) = \{ \underset{l}{\text{argmax}} \,\, \lmgen(\mathcal{P}_n, (o, l, i))\}.
\end{equation}
Given $N$ prompts, we then integrate the rewards in Eq.~\eqref{eq: temporal candidate} from each by employing the majority voting to establish temporally consistent rewards.
%Finally, we define temporally consistent reward $r_{\textnormal{B}}$ via majority voting on $R_T^j$:
\begin{equation}
\label{eq: temporal consistency}
    r^{T}(o, l | i) = \underset{r \in \hat{\mathbb{R}}}{\text{argmax}} \sum_{n=1}^N \mathbbm{1}_{(R_n^T(o, l | i) = r)}
\end{equation}

\subsection{A Domain-Grounded Reward Ensemble}
\label{sec: reward optimization}
From the spatio-temporally consistent rewards $r^C$, $r^S$, and $r^T$ calculated above, we derive domain-grounded rewards through their ensemble based on the alignment with given offline trajectories.  
We model unified rewards $\hat{r}$ as
\begin{equation}
\label{eq: unified reward}
\begin{array}{l}
    \mathbf{r}(o, l | i) = (r^C(o, l | i), r^S(o, l | i), r^T(o, l | i)) \\
    \mathbf{w}(o, l | i) = (w^C(o, l | i), w^S(o, l | i), w^T(o, l | i)) \\
    \hat{r}(o, l | i) = \langle \mathbf{r}(o, l | i), \mathbf{w}(o, l | i) \rangle
\end{array}
\end{equation}
where $\langle \cdot, \cdot \rangle$ is an inner product and  $w^C, w^S$ and $w^T$ are learnable weights. 
These $\mathbf{w}$ are generated by the reward orchestrator $\Psi_\theta$. 
It takes observation $o$, action $l$, and instruction $i$ as input, producing a softmax distribution for $\mathbf{w}$.
The orchestrator $\Psi_\theta$ is used to align the predicted return of a trajectory with the labeled return, i.e., the sparse reward $f_s(i, \tau)$:

\begin{equation}
\label{eq: orchestrator objective}
\resizebox{1.05\hsize}{!}{
$
    \begin{array}{l}
        \mathbf{w}(o_t, l_t | i) = \Psi_\theta(o_t, l_t, i) \\
        \mathcal{L}(\Psi_\theta) = \underset{\substack{(i, o_t, l_t) \sim \\ (i, \tau) \in \mathcal{D}}}{\mathbb{E}} \bigg[ \| \sum\limits_t \gamma^t \hat{r}(o_t, l_t | i) - \alpha f_s(i, \tau)\|^2 \bigg]
    \end{array}
$
    }
\end{equation}
where $\alpha$ is a hyperparameter.

Finally, using the augmented trajectory dataset that contains unified rewards $\hat{r}$ in Eq.~\eqref{eq: unified reward}, an agent can be trained via offline RL algorithms such as CQL~\cite{cql}. 
The two-staged reward estimation in $\oursol$ is outlined in Algorithm~\ref{alg: llm based off learning}.

\section{Experiments}
\subsection{Experiment Settings}
\label{subsec: experiment settings}
\noindent\textbf{Environment and dataset.}
For evaluation, we use VirtualHome (VH)~\cite{virtualhome}, a widely used realistic benchmark for household activities. 
VH features a diverse array of interactive objects (e.g., apples, couch) and basic behaviors (e.g., grasp, sit), enabling us to define 58 distinct actions for embodied agents.
We use 25 distinct tasks including activities such as sitting on a couch with several fruits, microwaving salmon, and organizing the bathroom counter.
To construct a training dataset $\mathcal{D}$ for offline RL, we begin with a single expert trajectory for each of these 25 tasks. We then augment each with random actions at intermediate steps that lead to failed trials. 
%incorporate trajectories that contain random actions at intermediate steps in each trajectory, resulting in failed trajectories. 
% --> 각 task별로 25개 이상의 많은 수의 failed trajectory가 있고, 각 task별로 1개의 failed trajectory를 sparse reward 0 으로 labeling 한 것입니다. 아래 설명이 정확히 맞는 것 같습니다.
For each expert trajectory, a sparse reward of $1$ is annotated to indicate success, while for sampled failed trajectories, a sparse reward of $0$ is annotated to denote failure. This follows Goal-POMDP representations used in long-horizon instruction-following tasks.
% for each of 25 failed trajectories augmented from the expert trajectory, a sparse reward of $0$ is annotated. 
%25 failed trajectories are randomly selected and labeled with a sparse reward of 0.

\begin{algorithm}[t]
    \caption{Two-staged $\oursol$}
    \begin{algorithmic}[1]
    \STATE Dataset $\mathcal{D}$, MDP-QA dataset $\mathcal{D}_{\textnormal{QA}}$
    \STATE Prompts $\mathcal{P}_1, \cdots, \mathcal{P}_N$ for LLM $\lmgen$
    \STATE Reward orchestrator $\Psi_\theta$
    \STATE Reward-augmented dataset $\bar{\mathcal{D}} = \varnothing$ 
    
    \nonumber \textit{\textbf{/* Spatio-Temp. Consistent Rewards */}}
    \FOR{$(i, (o, l, o')) \in (i, \tau) \in \mathcal{D}$}
    \STATE Reward-augmented trajectory $\bar{\tau} = \varnothing$
    \STATE $r^C \longleftarrow r^C(o, l | i)$ using Eq~\eqref{eq: majority consistency}
    \STATE $r^S \longleftarrow r^S(o, l | i)$ using $\mathcal{D}_{\textnormal{QA}}$ and Eq~\eqref{eq: structural candidate},~\eqref{eq: structural consistency}
    \STATE $r^T \longleftarrow r^T(o, l | i)$ using Eq~\eqref{eq: temporal candidate},~\eqref{eq: temporal consistency}
    \STATE $\bar{\tau} \longleftarrow \bar{\tau} \cup \{o, l, o', (r^C, r^S, r^T)\}$
    \IF{len($\tau$) = len($\hat{\tau}$)}
    \STATE $\bar{\mathcal{D}} \longleftarrow \bar{\mathcal{D}} \cup \{ (i, \bar{\tau}) \}$
    \ENDIF
    \ENDFOR
    
    \nonumber \textit{\textbf{/* Domain-Grounded Rewards in~\ref{sec: reward optimization} */}}
    \REPEAT
    \STATE Sample $(i, \bar{\tau}) \sim \bar{\mathcal{D}}$
    \STATE $\forall t$, compute $\mathbf{r}(o_t, l_t | i)$ using Eq~\eqref{eq: unified reward} 
    \STATE $\forall t$, compute $\mathbf{w}(o_t , l_t | i)$ using Eq~\eqref{eq: orchestrator objective}
    \STATE $\forall t$, $\hat{r}(o_t, l_t | i) \longleftarrow \langle \mathbf{r}(o_t, l_t | i), \mathbf{w}(o_t, l_t | i) \rangle$ 
    \STATE $\mathcal{L}(\Psi_\theta) \longleftarrow \| \sum\limits_{t} \gamma^t \hat{r}(o_t, l_t | i) - f_s(i, \tau)\| ^ 2$
    \STATE $\Psi_\theta \longleftarrow \Psi_\theta - \nabla_\theta \mathcal{L}(\Psi_\theta)$
    \UNTIL{converge}
    
    \end{algorithmic}
    \label{alg: llm based off learning}
\end{algorithm}

\noindent\textbf{Evaluation instruction.}
We employ two distinct instruction types to assess the agent's ability to handle different goal representations. 
A \textbf{Fine-grained} instruction type provides a detailed task description, often including specific actions performed to achieve certain goal conditions pertinent to the instruction-following task.
An \textbf{Abstract} instruction type provides a more abbreviated and generalized task description, focusing on broader objectives without detailing each action. 
Each of the 25 tasks is assessed using 5 fine-grained and 5 abstract instructions, resulting in a total of 250 distinct instructions being tested. These instructions have not been included within the offline training dataset. 

\noindent\textbf{Evaluation metrics.}
We use three metrics, consistent with previous works~\cite{progprompt, llmplanner}. 
\textbf{SR} measures the percentage of tasks successfully completed, defining success as the completion of all goal conditions for a task; 
\textbf{CGC} measures the percentage of completed goal conditions; 
\textbf{Plan} measures the percentage of the action sequence that continuously matches with the ground-truth sequence from the beginning.

\noindent\textbf{Baselines.}
We compare $\oursol$ with different categories of agents:
\textbf{RL agents}, in which an LLM is solely used for estimating rewards to train a separate RL agent, without directly using the LLM for online interaction;
\textbf{LM agents}, in which either a small language model (sLM) or LLM is used to directly interact with the environment as an online agent. These are in contrast to the RL agents that use LLMs solely for agent training.
In this LM agent category, to provide an evaluation under the compatible computational efficiency conditions with the RL agent category, we include \textbf{sLM-based agents} as well as \textbf{LLM-based agents}.

The \textbf{RL agent} category baselines include 
i) Lafite-RL~\cite{Lafite_RL}, which evaluates actions as good (1), neutral (0), or bad (-1) using an LLM, and integrates the evaluations with environmental rewards; 
ii) RDLM~\cite{RDLM}, which uses an LLM to evaluate trajectory returns using dynamically sampled in-context demonstrations; 
iii) Self-Consistency~\cite{self-consistency}, which generates multiple reward candidates via a single CoT prompt, taking a majority vote on them;
and iv) GCRL, which relies on given sparse rewards related to goal conditions.

The \textbf{LM agent} category baselines include 
v) SayCan~\cite{saycan}, which employs an offline dataset to learn the affordance scores combined with an LM's prediction; 
vi) LLM-Planner~\cite{llmplanner}, which uses an expert dataset for retrieval-augmented task planning; 
vii) ProgPrompt~\cite{progprompt}, which uses engineered programmatic assertion syntax to verify the pre-conditions of action execution.

% Each LM agent baseline is configured with both sLMs (GPT2-774M and 4-bit quantized LLaMA3-8B) and LLMs (Gemini 1.0\footnote{The parameter count of Gemini 1.0 has not been disclosed.} and LLaMA3-8B). 
Each LM agent baseline is configured with both sLMs (GPT2-774M and 4-bit quantized LLaMA3-8B) and LLMs (Gemini 1.0 Pro and LLaMA3-8B). 
The implementation of LM agent baselines with a larger LLaMA3-70B model can be found in Appendix~\ref{appendix subsec: llama3-70b for lm agents}.
% %
\color{black}

For our $\oursol$ and the \textbf{RL agent} category, we use Gemini 1.0 Pro for the reward estimator $\lmgen$ and adapt the GPT2-based model architecture having 117M parameters to implement the agent policies that learn from their respective rewards. We also employ the CQL~\cite{cql} offline RL algorithm in conjunction with the DDQN~\cite{ddqn} to handle the discrete action space in our environment. Details of the experiments are in Appendix.

\subsection{Main Results}
\noindent\textbf{Instruction-following task performance.}
Table~\ref{tab: single domain} presents a performance comparison of our $\oursol$ and the baselines from different categories, including RL agents, LLM-based agents, sLM-based agents across metrics such as \textbf{SR}, \textbf{CGC}, and \textbf{Plan}. 
%
% In the table, we also indicate the model sizes of the agent policy networks in parentheses, except for Gemini 1.0, whose number of parameters has not been disclosed.
\color{black}
\begin{itemize}[leftmargin=*]
    \item $\oursol$ outperforms all the RL agent baselines by a significant margin, achieving average gains of $20.0\%$, $15.2\%$, and $5.6\%$ over the most competitive RL agent baseline Self-Consistency in \textbf{SR}, \textbf{CGC}, and \textbf{Plan}, respectively.
    \item Furthermore, the performance of $\oursol$ is on par with the LLM-based agents, with only a slight performance drop compared to SayCan-Gemini and ProgPrompt-Gemini, while it surpasses the other LLM-based agents (i.e., all with LLaMA3 and LLM-Planner-Gemini). These results are especially noteworthy, considering the significantly different model sizes between $\oursol$ (GPT2-based-117M) and other LLM-based agents (i.e., Gemini, LLaMA-8B). These demonstrate $\oursol$'s ability to learn long-horizon instruction-following tasks within specific domains using minimal domain-specific knowledge, such as partially annotated rewards.
    \item We observe that the sLM-based agents using 4-bit quantized LLaMA3 (LLaMA3Q) and GPT2 exhibit lower performance than the others, including our $\oursol$, due to their dependency on the limited reasoning capabilities of sLMs. 
    \item Additionally, $\oursol$ demonstrates relatively robust performance across different instruction types compared to LLM-based agents. This can be attributed to $\oursol$'s ability to learn from a broad range of semantically similar instructions, which are generated by the LLM and included in the offline dataset. This enables the framework to better generalize to abstract instructions.
\end{itemize}

% caption 으로 모델 사이즈 넣어두기 (테이블 안에 X, Gemini 언급 하지않기)

\begin{table}[t]
    \centering
    \begin{adjustbox}{width=0.49 \textwidth}
    \begin{tabular}{l ccc ccc}
    \midrule
       & \multicolumn{3}{c}{\textbf{Fine-grained}} & \multicolumn{3}{c}{\textbf{Abstract}} \\
       % \cmidrule(lr){1-1}
       RL agent & \textbf{SR} & \textbf{CGC} & \textbf{Plan} & \textbf{SR} & \textbf{CGC} & \textbf{Plan} \\
       \cmidrule(lr){1-1} \cmidrule(lr){2-4} \cmidrule(lr){5-7}
       % \midrule
       \textbf{$\oursol$} &  66.4 & 74.5 & 69.5 &   57.6 & 68.3 &  64.8 \\
       \textbf{Lafite-RL} & 30.4  & 50.9  & 35.1 &  17.6  & 37.8  &  23.1  \\
       \textbf{RDLM} &  20.0 &  42.0 & 31.7 & 4.0  & 23.3  & 23.9  \\
       \textbf{Self-Consistency} & 43.2& 56.9 &  61.4 & 40.8 & 55.6 & 61.7 \\
       \textbf{GCRL} &  5.6 & 26.0 & 22.3 & 8.0 & 28.4 & 17.1 \\
       % \cmidrule(lr){1-1}
       \midrule
       LLM-based agent \\
       % \cmidrule(lr){1-1}
       \midrule
       \textbf{SayCan-Gemini} & 72.0 & 78.2 & 73.8 & 6.9  & 25.2  & 23.2  \\
       \textbf{SayCan-LLaMA3} & 4.8 & 22.4 & 63.8 & 3.2  &14.6 & 20.0 \\
       
       \textbf{ProgPrompt-Gemini} &  72.8 & 80.4 & 80.2 & 32.0 & 49.2 & 24.3 \\
       \textbf{ProgPrompt-LLaMA3} &  68.0 & 74.5 & 50.5 & 16.5 & 29.5 & 8.2  \\
       
       \textbf{LLM-Planner-Gemini} & 55.2 & 63.8 & 59.7 & 2.1 & 18.1 & 0.0 \\
       \textbf{LLM-Planner-LLaMA3} & 15.1 & 34.0 & 30.6 & 2.0 & 15.4 & 0.6 \\
       
       \midrule
       sLM-based agent \\
       % \cmidrule(lr){1-1}
       \midrule
       \textbf{SayCan-LLaMA3Q} & 4.8 & 21.6 & 62.6 & 0.0 & 15.4  & 0.4 \\
       \textbf{SayCan-GPT2} & 0.0 & 14.7 & 0.0 & 0.0 & 14.7  & 0.0  \\
       \textbf{ProgPrompt-LLaMA3Q} & 43.2 & 68.2 & 68.8 & 15.2 & 34.5 & 31.1 \\
      \textbf{ProgPrompt-GPT2} & 0.6 & 16.7 & 6.0 & 0.0 & 8.8 & 0.4\\
       \textbf{LLM-Planner-LLaMA3Q} & 12.4 & 31.1 & 8.9 & 0.6 & 13.9 & 0.2 \\
       \textbf{LLM-Planner-GPT2}& 0.0 & 12.6 & 0.0 & 0.0 & 12.6 & 0.0 \\
    \midrule
    \end{tabular}
    \end{adjustbox}
    % \caption{Instruction-following task performance in SR, CGC, and Plan metrics}
    % \caption{
    % Instruction-following task performance in SR, CGC, and Plan metrics. 
    % RL agent (117M), LLaMA3 (8B), and GPT2 (774M) in instruction-following task.}
    \caption{
Instruction-following task performance in SR, CGC, and Plan metrics. 
Agent policy model sizes: RL agents (117M), LLM-based agents (Gemini and LLaMA3-8B), sLM-based agents (GPT2-774M and 4bit-quantized LLaMA3-8B).
}
    
    \label{tab: single domain}
\end{table}
% 0921: GPT2사이즈 표기하고, Cross-domain에 지금 해놓은 것처럼 똑같이 이름 변경

\noindent\textbf{Cross-domain performance.}
Here, we extend our evaluation scenarios to include domain shifts in the environment; i.e., the locations of key objects related to the given instructions differ from those in the training dataset.  
Specifically, we sample a subset of trajectories from the training dataset $\mathcal{D}$ and relabel their sparse rewards $f_s(i, \tau)$ to reflect the altered object locations. 
While keeping the spatio-temporally consistent rewards unchanged, we then retrain the reward orchestrator in Eq.~\eqref{eq: orchestrator objective} using these newly labeled sparse rewards. 
This approach facilitates the generation of domain-specific unified rewards for RL without the need to recalculate the consistency-based rewards themselves through LLM inferences.
%%%%%
We also incorporate this newly labeled dataset for the LM agent category. 
For instance, LLM-Planner adapts to this new environment domain by using the trajectories, which are relabeled as success, as demonstrations for task planning. 
Since other RL agent baselines, except GCRL, lack the ability to utilize domain information represented as sparse rewards, they are evaluated with the same policy as in the single-domain experiments.
%
% Table~\ref{tab: cross domain} shows the performance under conditions of domain shifts. 

\begin{table}[t]
    \centering
    \begin{adjustbox}{width=0.49 \textwidth}
    \begin{tabular}{l ccc ccc}
    \midrule
       & \multicolumn{3}{c}{\textbf{Fine-grained}} & \multicolumn{3}{c}{\textbf{Abstract}} \\
       RL agent & \textbf{SR} & \textbf{CGC} & \textbf{Plan} & \textbf{SR} & \textbf{CGC} & \textbf{Plan} \\
       \cmidrule(lr){1-1} \cmidrule(lr){2-4} \cmidrule(lr){5-7}
       
       \textbf{$\oursol$} & 60.0 & 66.3 & 69.4 & 45.0 & 55.0 & 42.5 \\
       \textbf{Lafite-RL} &  2.5 & 12.5 & 10.6 & 0.0 & 12.5 & 25.2 \\
       \textbf{RDLM}      & 15.0 & 23.8 & 22.5 & 3.8 & 18.8 & 23.6 \\
       \textbf{Self-Consistency} & 35.4 & 47.9 & 54.7 & 31.3 & 45.8 & 51.6 \\
       \textbf{GCRL}      & 0.0 & 6.3 & 8.5 & 0.0 & 6.3 & 10.9 \\
       \midrule
       LLM-based agent\\
       \midrule
       \textbf{SayCan-Gemini} &  12.5 & 18.8 & 26.6 & 0.0 & 8.3 & 0.0 \\
       \textbf{SayCan-LLaMA3} & 5.0 & 21.3 & 1.9 & 0.0 & 10.0 & 1.3 \\
       
       \textbf{ProgPrompt-Gemini} &   25.0 & 31.3 &  23.4 & 14.6 & 32.3 & 1.6 \\
       \textbf{ProgPrompt-LLaMA3} & 25.0 & 32.3 & 22.6 & 8.3 &  24.0 &  3.1 \\
       
       \textbf{LLM-Planner-Gemini} & 45.8 & 55.2 & 67.7 & 0.0 & 13.7 & 0.0 \\
       \textbf{LLM-Planner-LLaMA3} &  6.3 & 20.2 & 16.8 & 0.0 &  40.6 & 0.0 \\
       
       \midrule 
       sLM-based agent\\
       \midrule
       \textbf{SayCan-LLaMA3-Q} &  8.3 &  21.9 & 1.5 & 0.0 &  12.5 & 1.9 \\
       \textbf{SayCan-GPT2} & 0.0 & 6.3 & 0.0 & 0.0 &  6.3 & 0.0 \\
       \textbf{ProgPrompt-LLaMA3-Q} & 7.5 & 15.5 & 18.3 & 0.0 & 31.3& 0.0 \\
       \textbf{ProgPrompt-GPT2} & 0.0 & 8.3 & 0.3 & 0.0 & 6.3 & 0.0\\
       \textbf{LLM-Planner-LLaMA3-Q} & 2.1 & 15.2& 9.9 &  0.0 & 13.5 & 0.0 \\
       \textbf{LLM-Planner-GPT2} & 0.0 &  6.3 & 0.0 & 0.0 & 6.3 & 0.0 \\
    \midrule
    \end{tabular}
    \end{adjustbox}
    \caption{Cross-domain performance}
    \label{tab: cross domain}
\end{table}

\begin{itemize}[leftmargin=*]
    \item For this cross-domain assessment, as shown in Table~\ref{tab: cross domain}, $\oursol$ outperforms all the RL agent baselines, showing a minimal drop compared to the results in the single-domain experiments (in Table~\ref{tab: single domain}). 
    %obtaining $24.6\%$ higher in \textbf{SR} than the most competitive baseline LLM-Planner-Gemini. 
    Upon domain shifts, $\oursol$'s two-staged process adjusts the reward estimates to align with the target domain by the second stage conducting the adaptive ensemble in Eq.~\eqref{eq: orchestrator objective}. 
    In contrast, the RL agent baselines, which rely solely on the rewards derived from the LLM's commonsense reasoning, exhibit a diminished ability to adapt to specific domains, showing large drops compared to the results in the single-domain experiments. 
    \item   %LLM-based agent 비교
    We also observe that the LLM-based agent baselines experience large degradation in this cross-domain assessment;
    %
    %SayCan, for instance, relies on the LLM's  knowledge, which is difficult to ground in a specific environment using limited target domain data, leading to suboptimal performance.
    e.g., LLM-Planner relies on the LLM's knowledge, which is difficult to ground in a specific environment using only a few examples, leading to suboptimal performance.  
    %$\oursol$ achieves robust performance compared to the LLM-based agent baselines, obtaining $24.6\%$ higher in \textbf{SR} than the most competitive LLM-Planner-Gemini. 
%Similarly, LLM-based agents struggle with cross-domain adaptation.
%
%SayCan, for instance, relies on the LLM's general knowledge, which is difficult to ground in a specific environment using limited target domain data, leading to suboptimal performance.
\end{itemize}

%
%In $\oursol$, the pre-estimated rewards can be effectively optimized into a single, domain-specific reward using partially labeled sparse rewards. 
%
%In contrast, other RL agent baselines, which rely solely on rewards derived from the LLM's commonsense reasoning, exhibit a diminished ability to adapt to specific domains.
%
%Similarly, LLM-based agents struggle with cross-domain adaptation.
%
%SayCan, for instance, relies on the LLM's general knowledge, which is difficult to ground in a specific environment using limited target domain data, leading to suboptimal performance.

% LLM-Planner, for instance, relies on the LLM's general knowledge, which is difficult to ground in a specific environment using only a few examples, leading to suboptimal performance.

\subsection{Ablation Studies}
% 교수님. 여기 Human reward와 비교하는 부분은 지우고, RL agent 성능만 비교해야 할 것 같아서요. 
% --> ok. 수정하면 알려주세요. 그 후에 수정할께요. 넵.
% In these ablation studies, we evaluate the agreement between LLM-based  rewards and human-labeled rewards, using the latter ones as a reference for comparison. 
% %
% The human-labeled rewards are obtained under necessary environmental contexts for the accomplishment of the instruction, such as the location of objects, which are provided to the human labelers.
% %
% We use three metrics for this comparison:
% %
% \textbf{Highest Reward Recall (HRR)} measures the recall of the LLM predicting the highest reward for actions that humans consider to have the highest reward.
% \textbf{Average Reward Recall (ARR)} extends this recall measurement to all reward levels beyond the highest rewards, taking the average recall across all levels.
% \textbf{Pairwise Reward Ordering (PRO)} evaluates the relative ordering of rewards rather than their absolute values. It measures the percentage of action pairs for which the LLM predicts the same reward ordering as the human labels.

\noindent\textbf{Spatio-temporally consistent rewards.}
To verify that the contextual, structural, and temporal consistencies (in Section~\ref{sec: consistent rewards}) effectively complement each other in LLM-based reward estimation, we test different combinations of these consistencies in the ensemble of rewards. Table~\ref{tab: ablation consistency} demonstrates that $\oursol$, which utilizes all three, consistently outperforms the others. This specifies that the combination of $\mathbf{w}$ and rewards derived from partial consistencies alone is limited in generating unified rewards that significantly benefit RL, while the ensemble weights $\mathbf{w}$ can be adjusted via Eq.~\eqref{eq: orchestrator objective}.

\begin{table}[h]
    \centering
    \begin{adjustbox}{width=0.49 \textwidth}
    \begin{tabular}{l ccc ccc}
    \midrule
       & \multicolumn{3}{c}{\textbf{Fine-grained}} & \multicolumn{3}{c}{\textbf{Abstract}} \\
       
       & \textbf{SR} & \textbf{CGC} & \textbf{Plan} & \textbf{SR} & \textbf{CGC} & \textbf{Plan} \\
       % \cmidrule(lr){2-4} \cmidrule(lr){5-10}
       \cmidrule(lr){2-4}  \cmidrule(lr){5-7}
       % \cmidrule(lr){8-10} 
       % T = T /R = C /P = S
       \textbf{$\oursol$} & 66.4 & 74.5 & 69.5 &   57.6 & 68.3 &  64.8\\
       \textbf{CS} & 64.8 &  67.9& 69.7 &  52.0 & 62.3 & 56.3 \\
       \textbf{ST} & 57.6 & 70.1 & 65.4 &  50.4 & 60.8 & 61.7  \\
       \textbf{CT} & 53.6 &  66.1 & 67.8 & 51.2 & 59.1 &  68.9 \\
       \textbf{C} & 47.2 & 58.6 & 59.7 & 47.1 & 57.1& 58.3\\
       \textbf{S} & 52.0 & 67.1 & 57.9 & 45.6 & 60.0 & 58.9 \\
       \textbf{T} & 45.6 & 57.8 & 55.7 & 41.6 &  51.2 & 50.8 \\
    \midrule
    \end{tabular}
    \end{adjustbox}
    \caption{Ablation on spatio-temporally consistent rewards. For example, CS denotes the use of contextual and structural consistencies, and T denotes the use of temporal consistency only, while $\oursol$ employs all three consistencies in the ensemble.}
    \label{tab: ablation consistency}
\end{table}

\noindent\textbf{Different LLMs for reward estimation.} % mod: raging -> ranging
To implement $\oursol$, which uses an LLM for offline reward estimation, we test a variety of LLMs ranging from open-source LLaMA3-8B to proprietary models GPT4 turbo, Gemini 1.0 Pro, and PaLM. 
In Table~\ref{tab: ablation llms}, we observe that LLaMA3-8B, which has significantly fewer parameters, does not achieve performance comparable to the proprietary models. 
%Among the proprietary models, PaLM exhibits the lowest performance, indicating that underlying design choices and training methodology of LLMs significantly influence their ability to reward estimation.
%
% mod: impact on -> impacts
Among the proprietary models, the more recent and advanced capable LLMs, such as GPT4 turbo and Gemini 1.0 Pro, demonstrate a strong ability in reward estimation that positively impacts agent offline learning. 

\begin{table}[h]
    \centering
    \begin{adjustbox}{width=0.49 \textwidth}
    \begin{tabular}{l ccc ccc}
    \midrule
       & \multicolumn{3}{c}{\textbf{Fine-grained}} & \multicolumn{3}{c}{\textbf{Abstract}} \\
       & \textbf{SR} & \textbf{CGC} & \textbf{Plan} & \textbf{SR} & \textbf{CGC} & \textbf{Plan} \\
       \cmidrule(lr){2-4} \cmidrule(lr){5-7} 
       % Open-Source \\
       % \midrule
       \textbf{LLaMA3} &  12.0 & 28.7 & 39.4 & 9.6 &  27.9 & 40.1 \\
       \textbf{PaLM} & 16.8 & 32.9 & 35.8 & 10.4 & 27.7 & 25.4 \\
       \textbf{GPT4 turbo} & 65.6 & 71.8 & 70.6 & 40.8 & 50.5  & 52.5\\
       \textbf{Gemini 1.0 Pro} & 66.4 & 74.5 & 69.5 &  57.6 & 68.3 &  64.8 \\
    \midrule
    \end{tabular}
    \end{adjustbox}
    \caption{Different LLMs for reward estimation}
    \label{tab: ablation llms}
\end{table}

\noindent\textbf{Reward ensemble scheme.}
We evaluate several approaches as alternatives to the reward ensemble scheme in Eq.~\eqref{eq: unified reward}.
First, we consider taking the \textbf{average} of rewards $r^C$, $r^S$, and $r^T$ to obtain a unified reward. Second, we employ a \textbf{majority voting} mechanism over the three rewards. As shown in Table~\ref{tab: ablation ensemble}, both Avg and Majority Voting result in degraded performance compared to $\oursol$. 
While the majority voting of spatio-temporally consistent rewards can provide a considerable degree of domain groundedness, $\oursol$ takes it a step further by employing the reward ensemble process using sparse rewards as guidance.
% but by aligning their ensemble with the sparse rewards, $\oursol$ can achieve maximum performance.
% This highlights the benefits of $\oursol$, which employs the two-staged ensemble process to integrate the three types of rewards using sparse rewards as guidance.
% 0615 coren 과 maj vote 차이가 적은 것에 대한 설명이 필요한 듯.

\begin{table}[h]
    \centering
    \begin{adjustbox}{width=0.49 \textwidth}
    \begin{tabular}{l ccc ccc ccc}
    \midrule
       & \multicolumn{3}{c}{\textbf{Fine-grained}} & \multicolumn{3}{c}{\textbf{Abstract}} \\
       
       & \textbf{SR} & \textbf{CGC} & \textbf{Plan} & \textbf{SR} & \textbf{CGC} & \textbf{Plan} \\
       \cmidrule(lr){2-4} \cmidrule(lr){5-7} \cmidrule(lr){8-10}
       % \textbf{ICL} & 00.0 & 00.0 & 00.0 & 17.2 & 39.8 & 21.8 & 12.0 & 33.7 & 16.5 \\
       % \midrule
       % Consistency \\
       % \midrule
       % \textbf{Temporal}& 37.24\% & 53.56\% & 54.12\% &  0.00\% & 0.00\% & 0.00\% \\
       % \textbf{Relational} &  0.00\% & 0.00\% & 0.00\% & 0.00\% & 0.00\% & 0.00\% \\
       % \textbf{Procedure} & 0.00\% & 0.00\% & 0.00\% & 0.00\% & 0.00\% & 0.00\% \\
       % \midrule
       % Ensemble \\
       % \midrule
       \textbf{$\oursol$} & 66.4 & 74.5 & 69.5 &  57.6 & 68.3 &  64.8  \\
       \textbf{Avg} & 53.6 & 63.7 & 55.6 &  43.2 & 55.2 &  57.7\\
       \textbf{Maj.Voting} & 60.8 &70.2 & 68.9 & 55.2& 62.7 & 63.7 \\ 
       % \textbf{Self-Con.} & 00.0 & 00.0 & 00.0 & 00.0 & 00.0 & 00.0 & 00.0 & 00.0 & 00.0 \\
       % \cline{1-2}
       % \midrule
       % LfDO \\
       % \midrule
       % \textbf{IL} & - & - & - & 9.6 & 26.4 & 14.9 & 0.0 & 0.0 & 0.0 \\
       % \textbf{IRL} & 0.00\% & 0.00\% & 0.00\% & 0.00\% & 0.00\% & 0.00\% \\
    \midrule
    \end{tabular}
    \end{adjustbox}
    \caption{Ablation on reward ensemble scheme}
    \label{tab: ablation ensemble}
\end{table}

\section{Related Works}
\noindent\textbf{LLMs for embodied environments.}
Leveraging LLMs as an instruction-following agent in embodied environments becomes a bedrock, capitalizing on LLM's reasoning capabilities~\cite{tree-planner, progprompt, LACMA, EMMA, yun2023emergence}. 
To overcome the limitation of LLMs' insufficient knowledge about specific domain conditions of the environment, prior works incorporate domain-related information.
\cite{saycan} utilizes an offline dataset to learn the value of actions, which is later combined with the LLM's token generation probability to calibrate the LLM's decision for different domains.
\cite{llm-planner} employs an expert dataset as a knowledge base for retrieval-augmented task planning.
%which results in environment-aware planning.
%
Unlike those directly employing LLMs as agent policies and requiring online LLM inferences, our study focuses on leveraging LLMs for reward estimation in offline RL, thus allowing for efficient agent structures.  

\noindent\textbf{LLMs for reward design.} % Embodied랑 상관 없이.
In RL, reward engineering is a long-standing challenge, traditionally tackled through manual trial-and-error or by leveraging domain knowledge from human experts. 
Inverse RL, on the other hand, aims to infer the underlying reward function from reward-free expert demonstrations~\cite{cooperative_irl, irl_classification}.
With the advent of capable foundation models, recent works have exploited them to produce reward functions~\cite{rlvlmf, guide_rl_llm, vlm_zeroshot_reward, vlm_reward}. 
\cite{RDLM} harnesses the in-context learning of LLMs to evaluate the episodes of high-level tasks.
\cite{EUREKA} leverages the code generation ability of LLMs, given environmental programming code, producing multiple code-based reward functions to train RL agents online and enhance them via feedback from agent training statistics.
Our $\oursol$ framework also leverages LLMs for reward design; however, the framework distinguishes itself by focusing on generating domain-grounded rewards without direct interaction with the environment, particularly in scenarios where the available information about the embodied environment is limited to sparse rewards.

\section{Conclusion}
We presented the reward ensemble framework $\oursol$ to achieve robust LLM-based reward estimation for offline RL, specifically tailored for embodied instruction-following tasks. The framework utilizes a spatio-temporal consistency-guided ensemble method for reward estimation. 
It generates multiple stepwise rewards on offline trajectories, with each reward focusing on a specific consistency related to contextual, structural, or temporal aspects, and then it integrates the multiple rewards into more domain-grounded ones via the sparse reward-aligned ensemble. As this work is the first to adopt LLMs for offline learning of embodied agents, we hope it can provide valuable insights into the development of LLM-driven training acceleration techniques. This is particularly significant for embodied agents involved in long-horizon instruction-following tasks, which are typically constrained by sparse reward signals.
%
%We hope our work provides valuable insights into the development of real-world instruction-following agents.

\section{Limitations}
Despite the robust performance achieved by $\oursol$, we identify that its success heavily depends on the capabilities of LLMs engaging in reward estimation, as shown by the ablation study in Table~\ref{tab: ablation llms}.
Our LLM-based reward estimation is conducted in an offline manner, i.e., without direct interaction with the environment.
%
% mod: pretrained 대신 pre-trained?
However, the dependency on the capabilities of an LLM can be problematic, especially when the target environment domain significantly differs from the pretrained knowledge of the LLM and the domain changes continuously over time after agent deployment.
%and the domain specification is only partially observable.
% when the target environment domain significantly differs from the pretrained knowledge of LLM. 
%
In these cases involving dynamic Goal-POMDP environments, the agent policy  learned offline by the dense rewards on the training dataset can degrade in terms of its task performance. The benefits of our ensemble method with the notion of spatio-temporal consistency are attributed to the effective alignment with the training dataset, and they can be limited in such non-stationary environment conditions. 
%RL through LLM inferences, even with the consideration of spatio-temporal consistency in conjunction with capable LLMs, may be insufficient to overcome this challenge.
%
We leave the exploration of methods to address this limitation as a direction for future work.

\section{Acknowledgements}
%%%%%%%%%%%%%%%%%%%%%%%%  RS (예전 번호)
% 0921: 사사 번호 받아서 고치기 -> 슬랙에 사사번호 올라오면 어떻게 써야되는지 민종, 우경, 원제님한테 물어보세요 okay
We would like to thank anonymous reviewers for their valuable comments and suggestions.
This work was supported
% IITP
by the Institute of Information \& Communications Technology Planning \& Evaluation (IITP) grant funded by the Korea government (MSIT) 
(RS-2022-II221045 (2022-0-01045),   % 자기주도
RS-2022-II220043 (2022-0-00043), % 개성형성
RS-2019-II190421 (2019-0-00421)% 인공지능  // 대학원 
),  
by the IITP-ITRC(Information Technology Research Center)(IITP-2024-RS-2024-00437633, 10\%) 
grant funded by MSIT, % ITRC
by the National Research Foundation of Korea (NRF) grant funded by MSIT (No. RS-2023-00213118), % NRF 신진
by BK21 FOUR Project (S-2024-0580-000), % BK21
and by Samsung Electronics. % samsung

\color{black}
% \section{Ethics Statement}

\bibliography{main}

\clearpage
\newpage

\appendix
\label{sec:appendix}

\section{Experiment Settings}
\subsection{Environment}
We utilize VirtualHome~\cite{virtualhome}, an environment and benchmark designed for simulating embodied household tasks. In this environment, actions related to household task activities are established by combining available manipulation behaviors and objects. These actions are executed sequentially to perform complex household tasks. $\oursol$ employs a configuration consisting of a house with 4 rooms, utilizing a total of 58 different actions. % manipulation behavior 
The actions are derived from the combinations of 8 distinct manipulation behaviors (find, grab, open, close, sit, put, put in, switch on) with various objects present within the environment.

\noindent\textbf{Single domain evaluation.}
For single domain experiments in Table~\ref{tab: single domain}, we evaluate each of 25 distinct tasks using a total of 10 instructions per task. These instructions are divided into two categories: 5 fine-grained instructions, which provide detailed descriptions of the task, and 5 abstract instructions, which offer a more general overview. Detailed examples of tasks used are presented in Table~\ref{table: vh_singledomain}.

\noindent\textbf{Cross domain evaluation.}
In the cross-domain setting, we assess tasks within an environment with altered object locations (e.g., relocating an apple from a desk to inside a refrigerator), as described in Table~\ref{table: cross-location}. 
We evaluate a total of 8 tasks from 25 tasks in the single domain evaluation, each with 5 instructions. This is due to the fact that several objects are unable to be relocated in a new layout. Similar to the single domain, each task is assessed with both fine-grained and abstract instructions, totaling 6 instructions per task. Detailed examples of tasks used in the cross-domain evaluation are presented in Table~\ref{table: vh_crossdomain}.

\subsection{Offline Dataset}
\label{appendix subsec: offline dataset}
To construct a training dataset $\mathcal{D}$ for offline RL, we use a single expert trajectory for each of the 25 distinct tasks. Each expert trajectory is augmented with random actions at intermediate steps that lead to failed trials. This process yields a total of approximately 8,000 trajectories for the offline dataset $\mathcal{D}$. For each expert trajectory, a sparse reward of $1$ is annotated to indicate its success, while for each sampled failed trajectory, a sparse reward of $0$ is annotated to denote its failure. Overall, we utilize one successful and one failed trajectory to establish the sparse rewards.
%

%%%% cross domain object위치 바꾼거 넣기

\section{Implementation}
In this section, we present the implementation details of our $\oursol$ and baselines. 
\subsection{$\oursol$ Implementation}
We implement our framework using Python v3.9.19 and the automatic gradient framework Jax v0.4.7.
The models are trained on a system with an NVIDIA RTX A6000 GPU.
The implementation details of $\oursol$ include these parts: (i) LLM-based reward estimation, (ii) spatio-temporal consistency consideration for estimated rewards, (iii) domain-grounded reward ensemble, and (iv) offline RL.

\subsubsection{LLM-based reward estimation}
\label{appendix subsec: llm-based reward estimation}
The LLM $\lmgen$ takes the user instruction $l$, observation $o$, and action $a$ as inputs, along with a prompt $\mathcal{P}$ so as to estimate the rewards for $a$ based on how they contribute to accomplishing $l$. We employ multiple $N$ prompts $\mathcal{P}_1, \cdots, \mathcal{P}_N$, which differ in their description methods for the reward estimation task, incorporation of in-context demonstrations, or use of chain-of-thought (CoT) prompts.
Specifically, we use 5 different types of prompts to create effective rewards: 
A naive prompt that includes the explanation of reward estimation tasks and required format, 
three in-context Learning (ICL) prompts that include distinct demonstrations,
and a CoT prompt that includes the human-written reasoning path of reward estimation.
Each prompt contains the rubric for the reward estimation, including which actions should receive which rewards. For example, a reward of 2 is given for an action that should follow, given the previously completed actions, and a reward of -1 is given for an action that involves searching for objects not related to the task.
The prompt examples are provided in Table~\ref{table: naiveprompt}, \ref{table: cotprompt}, \ref{table: iclprompt}, and~\ref{table: iclprompt2}.

In conjunction with the aforementioned prompts, we employ several LLMs: LLama-8B, Gemini 1.0 Pro, PaLM, and GPT4 Turbo. For GPT4 Turbo, the temperature of 0.5 is used, while the other models are set to the temperature of 0.7. The temperature setting is based on the characteristics of each model and aims to balance the trade-off between exploration and exploitation during the reward generation process. Table~\ref{tab: llm hyperparameters} specifies the LLMs used, their respective model sizes, and the temperature hyperparameters used to conduct the ablation study in Table~\ref{tab: ablation llms}.

\begin{table}[h]
    \centering
    \begin{adjustbox}{width=0.4 \textwidth}
    \begin{tabular}{l ccc ccc}
    \midrule

       \textbf{LLM} & \textbf{Model Size} & \textbf{Temperature}  \\
      \midrule
        LLaMA3 & 8B & 0.7 \\
        PaLM & - & 0.7 \\
        GPT4 turbo & - & 0.5 \\
        Gemini 1.0 Pro & - & 0.7   \\
    \midrule
    \end{tabular}
    \end{adjustbox}
    \caption{LLMs, their model sizes, and the temperature hyperparameters used in Table~\ref{tab: ablation llms}}
    \label{tab: llm hyperparameters}
\end{table}

\subsubsection{Spatio-Temporal Consistency}
Here, we provide the detailed description and mechanism of the spatio-temporal consistency including contextual, structural, and temporal ones, explained in Section~\ref{sec: consistent rewards}.

\paragraph{Contextual Consistency}
Contextual consistency involves estimating rewards using the previously introduced prompts and then applying the majority voting to the results.

\paragraph{Structural Consistency}
Structural consistency incorporates a process where the reward estimator $\lmgen$ self-checks its ability to reflect partial information about the environment through MDP-specific queries. 
To facilitate this, we generate an MDP-specific QA dataset $\mathcal{D}_{\textnormal{QA}} = \{q(o), a(o): o \in \tau \in \mathcal{D}\}$. The QA dataset consists of queries $q(o)$ that are easier to answer than the reasoning task of estimating rewards for actions, requiring only observation and instruction. By evaluating the correctness of the responses to these queries, we determine whether the reward estimation has been carried out while properly considering the internal structure of the environment.

To create the answers for the MDP-specific dataset $\mathcal{D}_{\textnormal{QA}}$, we employ GPT4 and use the queries that focus on identifying the objects that play a crucial role in achieving the given instruction. Through this process, we generate a total of 139 QA-pairs. Table~\ref{table: mdpqa_dataset} shows the examples of QA-pairs.

Given observation $o$, $\lmgen$ takes a query $q(o')$ along with a prompt $\mathcal{P}_n$ as input and generates a response $\lmgen(\mathcal{P}_n, q(o'))$. Here, $q(o')$ is chosen based on the sentence embedding similarity between $o$ and $o'$ using the sentence transformer model~\cite{sentence_transformer}.
We integrate the query $q(o)$ into the prompt $\mathcal{P}_n$ by directly appending it at the end of the prompt. Table~\ref{table: naiveprompt} shows the examples.
To determine how well the response aligns with the actual answer, we utilize a similarity-based evaluator $E$. Specifically, if the sentence embedding similarity between the response and the ground truth answer $a(o')$ is below a threshold of 0.5, the response is considered incorrect. 
% This is reflected in Eq.\eqref{eq: structural consistency} of the main manuscript.

\begin{table*}[h]
    \centering
    \footnotesize 
    \begin{tabularx}
        {\textwidth}
        {>{\raggedright\arraybackslash}X}
        \toprule
        \textbf{Question-Answer Pairs in the MDP-specific dataset} \\
        \midrule
        \textbf{Query 1:}\\
        Instruction: <instruction>\\
        Visible objects: paper, wallshelf, cereal, mouse, mug, creamybuns, crackers\\
        Among the currently visible objects, which objects are relevant to the task?\\
        \textbf{Answer 1}: \\wallshelf, cereal
        \\~\\
        \textbf{Query 2:}\\
        Instruction: <instruction>\\
        Visible objects: paper, cpuscreen, desk, keyboard, mouse, mug\\
        Among the currently visible objects, which objects are relevant to the task?\\
        \textbf{Answer 2}: \\ desk, cat \\
    \bottomrule
    \end{tabularx}
    \caption{MDP-specific dataset $\mathcal{D}_{\textnormal{QA}}$}
    \label{table: mdpqa_dataset}
\end{table*}

\paragraph{Temporal consistency}
Temporal consistency involves calculating the sequence of high-value actions $H_n(\tau)$ for each prompt $\mathcal{P}_n$:
\begin{equation}
    H_n(\tau) = \{ \underset{l}{\text{argmax}} \,\, \lmgen(\mathcal{P}_n, (o, l, i))\},
\end{equation}
where $o$ is an observation, $l$ is an action, and $i$ is an instruction.
Note that there are multiple sequences of high-value actions.
For each sequence of high-value actions in $H_n(\tau)$, we present a query $q(i, \tau, n)$ to $\lmgen$ to determine whether the sequence can accomplish the instruction $i$.
Table~\ref{table: temporalpropmt} provides an example of an actual query.
If the query is violated, i.e., $\lmgen(q(i, \tau, n))$ returns False, the reward for the action $l \in H_n(\tau)$ is disregarded.
Otherwise, if $l\notin H_n(\tau)$ or $\lmgen(q(i, \tau, n))$ returns True, the estimated reward $\lmgen(\mathcal{P}_n, (o, l, i))$ is included in the majority voting process to construct the temporally consistent reward.

An example illustrating how reward estimation changes according to contextual, structural, and temporal consistency can be found in Table~\ref{table: consistency_rewards1},~\ref{table: consistency_rewards2}, and~\ref{table: consistency_rewards3}.

%%% Temporal Consistent Promp %%%
\begin{table*}[h]
    \centering
    \footnotesize 
    \begin{tabularx}{\textwidth}{>{\raggedright\arraybackslash}X}
        \toprule
        \textbf{Backward-Verification Prompt} \\
        \midrule
        \textbf{Prompt 1:}\\
         Action List: [action list]\\
        From the list of actions provided above, I selected a few actions to form an action sequence like <action sequence>. If this sequence of actions is executed in order, is it possible to achieve <instruction>?\\
        Answer with only "possible" or "impossible."
        \\~\\
        \textbf{Prompt 2:}\\
        You have created a sequence of actions from the list above as  <action sequence> to achieve <instruction>.\\
        However, this sequence is incorrect because a subsequent action cannot be performed without the prior action being executed.\\
        State the number of steps that are in the wrong order. Only output the number. If there are multiple numbers, separate them with a comma.\\
    \bottomrule
        \end{tabularx}
    \caption{Prompt for backward-verification of temporal consistency}
    \label{table: temporalpropmt}
\end{table*}

\subsubsection{Domain-grounded Reward Ensemble}
To learn the ensemble method for the spatio-temporally consistent rewards $r^C, r^S$, and $r^T$, we train a reward orchestrator $\Psi_\theta$.

\begin{table}[h]
    \centering
    \begin{adjustbox}{width=0.4\textwidth}
    \begin{tabular}{l c}
    \midrule
       \textbf{Hyperparameters} & \textbf{Values} \\
      \cmidrule(lr){1-1} \cmidrule(lr){2-2}
          Network architecture & Bert for 3 classification \\
          batch size & 16 \\
          Activation function	& ReLU, Softmax \\
          learning rate & 1e-4  \\
          Gradient clipping & 3 \\
    \midrule
    \end{tabular}
    \end{adjustbox}
    \caption{Hyperparameters for reward orchestrator $\Psi_\theta$}
    \label{tab: hprm_orchestrator}
\end{table}

The orchestrator is responsible for aligning the trajectory's return, which is the accumulation of stepwise rewards $\hat{r}$, with the sparse reward $f_s(i, \tau)$ annotated on the trajectory, as explained in Eq.\eqref{eq: orchestrator objective}.
The trajectory's return is defined as the summation of rewards $\hat{r}$. However, the scale of the return varies depending on the length of the trajectory $H$. To align the return with the sparse reward $f_s(i, \tau) \in {-1, 1}$, proper normalization is needed.
Assuming that the LLM reward estimation $\lmgen(\mathcal{P}_n, (o, l, i))$ can take values within the range $[-K, K]$, we normalize the return by dividing it by $HK$. 
This normalization ensures that the return falls between -1 and 1, making it compatible with the sparse reward. The orchestrator is implemented using a Bert-based architecture~\cite{bert} adapted for a 3-class classification task. The hyperparameter settings for $\Psi_\theta$ are summarized in Table~\ref{tab: hprm_orchestrator}.

\subsubsection{Offline Reinforcement Learning }
Regarding the model structure of agent policy $\pi$, we adapt the GPT2 architecture with 58 heads to represent the action value. To optimize $\pi$, we use the Double DQN (DDQN) algorithm~\cite{ddqn} to handle the discrete action space in our environment, and also adopt Conservative Q-Learning (CQL)~\cite{cql} to address the q-value overestimation problem inherent in offline RL. The hyperparameter settings for $\pi$ are summarized in Table~\ref{tab: hprm_policy}.
\begin{table}[h]
    \centering
    \begin{adjustbox}{width=0.3\textwidth}
    \begin{tabular}{l c}
    \midrule
       \textbf{Hyperparameters} & \textbf{Values} \\
      \cmidrule(lr){1-1} \cmidrule(lr){2-2} 
          Network architecture & GPT2 \\
          Number of positions & 1536 \\
          Number of layers & 2 \\
          Number of heads & 4 \\
          Activation function & ReLU \\
          Residual dropout & 0.1 \\
          Embedding dropout & 0.1 \\
          Attention dropout & 0.1 \\
          Layer norm epsilon & 0 \\
          Embedding dimension & 768 \\
          Learning rate & 1e-4 \\
          Target update interval & 250 \\
          Discount factor $\gamma$ & 0.99 \\
          $\tau$ for soft target update & 0.005 \\
    \midrule
    \end{tabular}
    \end{adjustbox}
    \caption{Hyperparameters for policy $\pi$}
    \label{tab: hprm_policy}
\end{table}

\subsection{Baseline Implementation}

\subsubsection{RL Agents}
\noindent\textbf{Lafite-RL.} % mod: given offline dataset -> the given offline dataset
In Lafite-RL~\cite{Lafite_RL}, an LLM is utilized to estimate the reward of each action in a given offline dataset based on observation and instruction. The estimated reward is one of three values: good (1), neutral (0), or bad (-1). This intrinsic stepwise reward for each action is combined with the sparse reward within the given offline dataset to establish a reward augmented dataset for RL. We use the prompt in Table~\ref{table: naiveprompt} for LLM inferences. 
The agent policy structure and its training hyperparameters are the same as those used in $\oursol$.

\noindent\textbf{RDLM.}
In RDLM~\cite{RDLM}, an LLM is utilized to estimate the trajectory returns based on a description of the task and user-specified in-context demonstrations. 
%
% 0616 여기서 this 가 무엇인가요? --> (continually constructed) prompt 입니다.
% mod  difference by ->  difference in
While the prompts are continually constructed based on the agent's successful rollout in the original RDLM work, due to the differences in our offline learning setting, we implement the prompts through retrieval-augmented generation (RAG) for reward estimation.
In doing so, we manually establish a dataset of reward estimations based on a rubric, i.e., a scoring guideline for the estimation, which can be found in Table~\ref{table: naiveprompt}. 
We then dynamically retrieve three in-context demonstrations, considering the cosine similarity between the instruction, action execution history, and observation.
% 0616 여기의 prompt 는 위의 prompt 와는 다른 것인 것 같은데..? 설명이 좀 이상함. 
% We use the prompt in Table~\ref{table: iclprompt2} for this retrieval-augmented LLM inferences. 
The retrieved demonstrations are combined with the prompt in Table~\ref{table: iclprompt2}, which is then used for the retrieval-augmented LLM inference.
The agent policy structure and its training hyperparameters are the same as those used in $\oursol$.

% 0616 citation?
\noindent\textbf{Self-Consistency.}
In this baseline~\cite{self-consistency}, an LLM is queried to estimate rewards with a CoT prompt. 
The LLM samples multiple $K$ reward estimates, each based on different reasoning paths, and selects the most consistent answer.
In our implementation, we set $K=3$. The agent policy structure and its training hyperparameters are the same as those used in $\oursol$.

\subsubsection{LM Agents}
\noindent\textbf{Saycan.}
SayCan~\cite{saycan} utilizes a combination of an LLM planner and an affordance value function to generate feasible action plans based on given instructions. 
The LLM planner identifies suitable actions, while the affordance score for each action is computed using a pre-trained affordance function. This affordance score is integrated into the LLM's token generation probability to select the feasible action to accomplish the task.

In our implementation, we follow the approach used by the authors of LLM-Planner, which involves retrieval-augmented task planning based on expert trajectories. 
Also, given the challenges of training low-level policies in VirtualHome, we employ the LLM-Planner's strategy of providing SayCan with object data to define the value function. This approach grants SayCan to narrow down the list of potential actions the LLM needs to consider. 
This streamlines the decision-making process for the planner, enhancing its ability to select executable actions and effectively complete tasks.

\noindent\textbf{ProgPrompt.}
ProgPrompt~\cite{progprompt} uses a programming assertion syntax to verify the pre-conditions for executing actions and addresses failures by initiating predefined recovery actions.

We employ the same plan templates as ProgPrompt, which feature a Pythonic style where the task name is designated as a function, available actions are included through headers, accessible objects are specified in variables, and each action is delineated as a line of code within the function. We use dynamically sampled in-context demonstrations for the LLM Planner and provide ProgPrompt with oracle pre-conditions for each action.

\noindent\textbf{LLM-Planner.}
The LLM-Planner~\cite{llmplanner} employs templatized actions, k-nearest neighbors (kNN) retrievers, and an LLM planner. The action candidates for planning are established based on templates, which are combined with the objects visible in the environment. 
The LLM-Planner retrieves in-context examples from the expert trajectories within our offline dataset, utilizing kNN retrievers, which are then prompted to the LLM planner. Subsequently, the planner merges these action templates with currently visible objects to determine the action that is both achievable and capable of completing the task.
% We use the open-source code from the LLM-planner official github. 

\section{Modality for reward estimation.}
We also investigate the use of large multi-modal models (LMMs) for reward estimation. 
Unlike $\oursol$, which uses detected object names within the scene to represent the observation, LMMs can directly utilize image observations.
Table~\ref{tab: ablation modality} shows that the agent trained with LMM-estimated rewards exhibit lower performance compared to their LLM counterparts. In this test, we use different ensemble approaches, as described previously in Table~\ref{tab: ablation ensemble}. 
% \color{black}
%
We speculate that while the image itself can implicitly convey detailed environmental information, LMMs' limited representation capabilities in embodied environments may not be well-suited for high-level reasoning tasks such as reward estimation.

\begin{table}[h]
    \centering
    \begin{adjustbox}{width=0.48 \textwidth}
    \begin{tabular}{l ccc ccc}
    \midrule
       & \multicolumn{3}{c}{\textbf{LLM}} & \multicolumn{3}{c}{\textbf{LMM}} \\
       & \textbf{SR} & \textbf{CGC} & \textbf{Plan} & \textbf{SR} & \textbf{CGC} & \textbf{Plan} \\
       \cmidrule(lr){2-4} \cmidrule(lr){5-7}
       \textbf{$\oursol$} &  62.0 & 71.4 & 67.2 & 26.0 & 42.1 & 31.1 \\

       \textbf{Averaged} & 48.4 & 59.5 & 56.7 &   22.4 & 43.2 & 44.9 \\
       \textbf{Maj. Voting} & 58.0 & 66.45 & 66.3 &  23.2 & 41.1 & 41.6 \\ 
       % \textbf{Self-Con.} & 00.0 & 00.0 & 00.0 & 00.0 & 00.0 & 00.0 \\
    \midrule
    \end{tabular}
    \end{adjustbox}
    \caption{Modality for reward estimation}
    \label{tab: ablation modality}
\end{table}

\section{Additional Experiments}
\subsection{LLaMA3-70B for LM Agents}
\label{appendix subsec: llama3-70b for lm agents}
% 0923: Table12에 다른 LLM으로 실험한 SayCan, ProgPrompt, LLM Planner 결과도 들어가는게 맞을 듯. 그러니깐, openreview에 있는 표 그대로 쓰는게 맞는 듯하네요. ok
Here, we implement LM agents (i.e., SayCan, ProgPrompt, and LLM-Planner) with LLaMA3-70B, a highly capable LLM. Table~\ref{tab: llama3-70b} shows the single-domain performance with fine-grained instruction on VirtualHome, achieved by our CoREN and LLaMA3-70B-based LLM-agent baselines. We observe that leveraging LLaMA3-70B results in performance improvements for both ProgPrompt and SayCan, with ProgPrompt exhibiting particularly substantial gains. Specifically, ProgPrompt achieved an average increase of 8.8\% in success rate (SR) when using LLaMA3-70B as an online agent compared to using LLaMA3-8B or Gemini. We hypothesize that this improvement is not only due to the larger model size enhancing ICL performance but also due to LLaMA3's significantly superior code analysis capabilities compared to Gemini~\cite{gemini} . This advantage particularly benefits ProgPrompt's performance with its programmatic prompts. For LLM-Planner, Gemini was found to be more suitable than LLaMA3.
More importantly, our framework which uses LLM (Gemini 1.0 Pro) only during training, still demonstrates competitive performance compared to LLM-based agent baselines which use LLaMA3-70B as an embodied agent.
% 0921: 여기 table 추가 및 인용 참조하기

\begin{table}[h]
    \centering
    \begin{adjustbox}{width=0.48\textwidth}
    \begin{tabular}{lccc}
    \toprule
       & \multicolumn{3}{c}{\textbf{Fine-grained}} \\
       RL agent & \textbf{SR} & \textbf{CGC} & \textbf{Plan} \\
       \cmidrule(lr){1-4}
       \textbf{CoREN} & 66.4 & 74.5 & 69.5 \\
       \midrule
       LLM-based agent \\
       \midrule
      \textbf{SayCan-Gemini} & 72.0 & 78.2 & 73.8 \\
       \textbf{SayCan-LLaMA3-70B} & 73.6 & 77.2 & 70.2 \\
       \textbf{SayCan-LLaMA3-8B} & 4.8 & 22.4 & 63.8 \\
       \textbf{ProgPrompt-Gemini} & 72.8 & 80.4 & 80.2 \\
       \textbf{ProgPrompt-LLaMA3-70B} & 79.2 & 83.9 & 74.4 \\
       \textbf{ProgPrompt-LLaMA3-8B} & 68.0 & 74.5 & 50.5\\
       \textbf{LLM-Planner-Gemini} & 55.2 & 63.8 & 59.7\\
       \textbf{LLM-Planner-LLaMA3-70B} & 39.2 & 53.2 & 51.4 \\
       \textbf{LLM-Planner-LLaMA3-8B} & 15.1 & 34.0 & 30.6 \\
    \bottomrule
    \end{tabular}
    \end{adjustbox}
    \caption{Single-domain performance with fine-grained instruction on VirtualHome using LLaMA3-70B}
    \label{tab: llama3-70b}
\end{table}

\subsection{Experiments on ALFRED Environment}
To verify the generalization capability of our framework, we conduct additional experiments in the ALFRED environment~\cite{ALFRED}.

\begin{table}[t]
    \centering
    \begin{adjustbox}{width=0.49 \textwidth}
    \begin{tabular}{l ccc ccc}
    \midrule
       & \multicolumn{3}{c}{\textbf{Fine-grained}} & \multicolumn{3}{c}{\textbf{Abstract}} \\
       % \cmidrule(lr){1-1}
       RL agent & \textbf{SR} & \textbf{CGC} & \textbf{Plan} & \textbf{SR} & \textbf{CGC} & \textbf{Plan} \\
       \cmidrule(lr){1-1} \cmidrule(lr){2-4} \cmidrule(lr){5-7}
       % \midrule
       
       \textbf{$\oursol$ } &  72.00 & 84.2 & 79.4 &  56.8 & 71.73 &  70.5 \\
       \textbf{Lafite-RL} & 8.0  & 17.6  & 38.4 &  39.2  & 55.8  &  72.1  \\
       \textbf{RDLM} &  46.4  & 61.7  & 72.4 &  14.4  & 23.8  &  40.7  \\
       \textbf{Self-Consistency} & 32.8 &  40.6 & 52.7 & 30.4 & 37.4  & 53.1  \\
       \textbf{GCRL} &  5.6 & 12.6 & 15.2 & 5.2 & 12.8 & 16.7 \\
       % \cmidrule(lr){1-1}
       \midrule
       LLM-based agent \\
       % \cmidrule(lr){1-1}
       \midrule
       \textbf{SayCan-Gemini} & 81.6 & 85.0 & 73.8 & 52.4  & 52.4  & 58.8  \\
       \textbf{SayCan-LLaMA3} &  70.4 & 75.7 & 70.3 & 39.2  & 40.0 & 48.5 \\
       
       \textbf{ProgPrompt-Gemini} &  68.8 & 78.0 & 40.6 & 48.0 & 57.4 & 27.9 \\
       \textbf{ProgPrompt-LLaMA3} & 70.4 & 78.2 & 76.0 & 32.0  & 47.2 & 22.6 \\
       
       \textbf{LLM-Planner-Gemini} & 44.4 & 51.8 & 58.7 & 15.7 & 24.6 & 0.0 \\
       \textbf{LLM-Planner-LLaMA3} & 16.0 & 27.5 & 46.3 & 6.4 & 13.3 & 34.1 \\
       
       \midrule
       sLM-based agent \\
       % \cmidrule(lr){1-1}
       \midrule
       \textbf{SayCan-LLaMA3Q} & 74.4 & 79.1 & 75.5 & 36.8 & 37.6  & 45.0\\
       \textbf{SayCan-GPT2} & 0.0 & 8.0 & 0.8 & 0.0 & 8.0  & 1.2 \\
       \textbf{ProgPrompt-LLaMA3Q} & 69.6 & 78.5 & 40.3 & 30.4 & 46.9 & 23.6 \\
      \textbf{ProgPrompt-GPT2} &  12.0 & 25.3 & 22.4 & 14.4 & 30.9 & 18.6 \\
       \textbf{LLM-Planner-LLaMA3Q} & 8.8 & 17.2 & 40.3 & 2.4 & 10.9 & 32.4 \\
       \textbf{LLM-Planner-GPT2} & 1.6 & 9.0 & 32.4 & 1.4 & 8.4 & 32.2\\
    \midrule
    \end{tabular}
    \end{adjustbox}
    \caption{Instruction-following task performance in SR, CGC, and Plan metrics in ALFRED}
    \label{tab: alfred single domain}
\end{table}

While both ALFRED and VirtualHome simulate household activities, they exhibit distinct characteristics. In VirtualHome, agents have access to broader environmental information, allowing immediate execution of actions like "find refrigerator" due to pre-encoded location data. Conversely, in ALFRED, executing such actions requires preliminary low-level actions like "go to kitchen," as the agent must navigate based on its understanding of the environment's spatial layout.

These differences introduce additional challenges when using LLMs as reward estimators in ALFRED, making it more difficult to generate rewards well-grounded in the environment's domain. This increased complexity provides a more rigorous test of our framework's ability to generate accurate rewards. Furthermore, the ALFRED environment offers a pre-existing offline dataset, making it suitable to verify our work's targeted contribution of LLM-based offline reward estimation based on a given dataset.

Table~\ref{tab: alfred single domain} compares the single-domain performance of $\oursol$ and baselines, while Table Table~\ref{tab: alfred cross domain} shows their cross-domain performance. The results demonstrate that our CoREN maintains state-of-the-art performance compared to other RL agent category baselines and shows comparable performance to LM agent category baselines. Specifically, CoREN outperforms RL agent baselines by a significant margin, achieving average gains of 14.4\% in Success Rate (SR) over the most competitive RL baseline, RDLM. 

For the evaluation in ALFRED, we utilize expert trajectories from the ALFRED benchmark. We then augment these long-horizon trajectories by appending specific actions at intermediate steps of the trajectories. This process resulted in 25 distinct tasks, each defined by a distinct sequence of actions. Using these 25 expert trajectories, we followed the same offline dataset construction process as outlined in Section~\ref{subsec: experiment settings} and~\ref{appendix subsec: offline dataset}. Also, each of the 25 tasks is evaluated using 5 fine-grained and 5 abstract instructions, resulting in a total of 250 test instructions. We use the same prompts for LLM-based reward estimation as those used in VirtualHome.
% 0921: 여기 table 두 개 추가 및 인용 참조하기

\begin{table}[t]
    \centering
    \begin{adjustbox}{width=0.49 \textwidth}
    \begin{tabular}{l ccc ccc}
    \midrule
       & \multicolumn{3}{c}{\textbf{Fine-grained}} & \multicolumn{3}{c}{\textbf{Abstract}} \\
       % \cmidrule(lr){1-1}
       RL agent & \textbf{SR} & \textbf{CGC} & \textbf{Plan} & \textbf{SR} & \textbf{CGC} & \textbf{Plan} \\
       \cmidrule(lr){1-1} \cmidrule(lr){2-4} \cmidrule(lr){5-7}
       % \midrule
       
       \textbf{$\oursol$} & 66.7  & 72.2 & 67.0 &  62.2 & 71.7 & 72.2 \\
       \textbf{Lafite-RL} & 11.1  & 25.6 & 29.6 & 15.6  & 35.8 & 33.1  \\
       \textbf{RDLM} &  48.0 & 58.0  & 51.3 &  15.6  & 22.2  &  22.4  \\
       \textbf{Self-Consistency} & 0.0 &  11.1 & 7.4 & 0.0 &  11.1 & 7.4  \\
       \textbf{GCRL} &  2.2 & 7.8 &  3.0 & 0.0 & 5.6 & 4.6\\
       % \cmidrule(lr){1-1}
       \midrule
       LLM-based agent \\
       % \cmidrule(lr){1-1}
       \midrule
       \textbf{SayCan-Gemini} & 44.4 & 50.0 & 51.2 & 0.0  & 15.5 & 25.9  \\
       \textbf{SayCan-LLaMA3} & 40.0 & 45.5 & 46.1 & 11.1 & 22.2 & 29.2 \\
       
       \textbf{ProgPrompt-Gemini} & 33.3 & 44.4 & 42.5 & 0.0 & 11.1 & 0.0 \\
       \textbf{ProgPrompt-LLaMA3} & 31.1 & 42.2 & 39.0	& 4.4 & 13.3 & 0.0\\
       
       \textbf{LLM-Planner-Gemini} & 44.4 & 50.0 & 51.2 & 11.1 & 24.4 & 20.7\\
       \textbf{LLM-Planner-LLaMA3} & 26.6 & 34.4 & 44.4 & 0.0 & 14.4 & 25.9\\
       
       \midrule
       sLM-based agent \\
       % \cmidrule(lr){1-1}
       \midrule
       \textbf{SayCan-LLaMA3Q} & 44.4 & 50.0 & 50.3 & 2.2 & 17.7 & 27.9 \\
       \textbf{SayCan-GPT2 } & 0.04 & 11.14 & 0.74 & 0.04 & 11.14 & 0.7\\
       \textbf{ProgPrompt-LLaMA3Q} & 26.6 & 37.7 & 33.8 & 0.0 & 10.0 & 0.0 \\
      \textbf{ProgPrompt-GPT2} & 0.0 & 11.1 & 1.8 & 0.0 & 11.1 & 0.0\\
       \textbf{LLM-Planner-LLaMA3Q} & 20.0 & 28.8 & 46.6 & 0.0 & 15.5 & 29.4\\
       \textbf{LLM-Planner-GPT2} & 4.4 & 13.3 & 16.6 & 0.0 & 14.4 & 32.6\\
    \midrule
    \end{tabular}
    \end{adjustbox}
    \caption{Cross-domain performance in ALFRED}
    \label{tab: alfred cross domain}
\end{table}

% mod to investicate  -> to investigate
\subsection{The Number of Prompts}
To investigate the impact of the number of prompts used in calculating spatio-temporally consistent rewards, we vary the number of prompts used to compute contextually, structurally, and temporally consistent rewards in Equations~\eqref{eq: majority consistency},~\eqref{eq: structural consistency}, and~\eqref{eq: temporal consistency}.

As explained in Section~\ref{appendix subsec: llm-based reward estimation}, we use 5 different types of prompts in a main manuscript: a naive prompt explaining the reward estimation task and required format, three in-context learning (ICL) prompts with distinct demonstrations, and a Chain-of-Thought (CoT) prompt including a human-written reasoning path for reward estimation.

Here, we explored 1, 3, 5, and 7 prompts for LM-based reward estimation:
\begin{itemize}[leftmargin=*]
    \item 1 prompt: CoT prompt only
    \item 3 prompts: 2 ICL prompts and 1 CoT prompt
    \item 5 prompts: As in the main manuscript
    \item 7 prompts: Added 2 distinct ICL prompts with new demonstrations
\end{itemize}

As shown in Table~\ref{tab:num of prompt}, we observe a positive correlation between the number of prompts and agent performance. This demonstrates that increasing the diversity of prompts enhances the robustness of our reward estimation process. The improved performance with more prompts suggests that our framework effectively leverages multiple perspectives to generate more accurate and consistent rewards.

% 0921: 여기 TABLE 추가 및 인용  
\begin{table}[h]
    \centering
    \begin{adjustbox}{width=0.3\textwidth} 
    \begin{tabular}{cc}
    \toprule
       \multirow{2}{*}{Num. Prompt} & Fine-grained \\
        & SR \ \ CGC \ \ Plan \\  
    \midrule
       1 & 36.8 \ \ 49.5 \ \ 56.7 \\
       3 & 50.4 \ \ 61.5 \ \ 65.2 \\ 
       5 & 66.4 \ \ 74.5 \ \ 69.5 \\ 
       7 & 69.6 \ \ 74.0 \ \ 76.9 \\ 
    \bottomrule 
    \end{tabular}
    \end{adjustbox}
    \caption{Performance Based on the Number of Prompts}
    \label{tab:num of prompt}
\end{table}

%% single domain taks examples
\begin{table*}[h]
\begin{center}
\Large
\resizebox{\textwidth}{!}{
\begin{tabular}{lc c c c cc}
\toprule
 \multirow{2}{*}{ID} & \multirow{2}{*}{Visual Observation} & \multirow{2}{*}{Goal}  & \multicolumn{2}{c}{Instruction Information} \\
 \cmidrule(rl){4-5}
 & & & Fine-grained & Abstract \\
\midrule

\multirow{6}{*}{1} & \multirow{6}{*}{\includegraphics[scale=0.18]{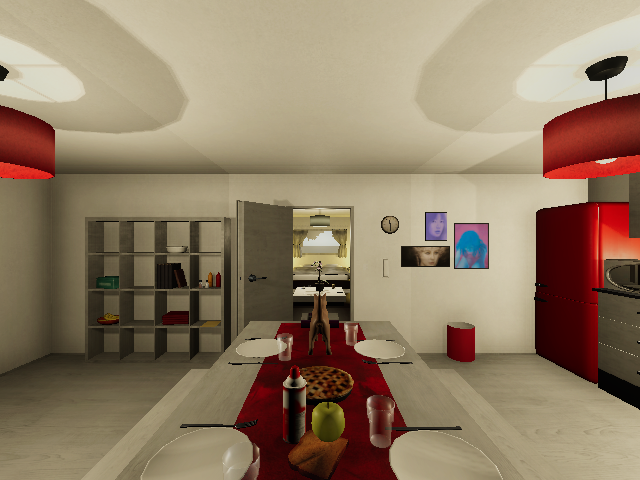}}  & \multirow{6}{*}{\begin{tabular}{c} {Apple on the kitchen table} \\ {Bread slice on the kitchen table} \\ {Apple on a bread slice} \end{tabular}}
& \multirow{6}{*}{\begin{tabular}{c} {Retrieve the apple from the coffee table, } \\ {walk to the toaster,} \\ { grab a bread slice,} \\
{ go to the kitchen table, } \\
{set the bread slice on the kitchen table, } \\
{put the apple on the bread slice.} \end{tabular}}& \multirow{5}{*}{\begin{tabular}{c} {Experience the natural crispness of} \\{apples in a tasty sandwich.} \end{tabular}} \\
  \\
  \\
  \\
  \\
  \\
\multirow{5}{*}{2} & \multirow{5}{*}{\includegraphics[scale=0.18]{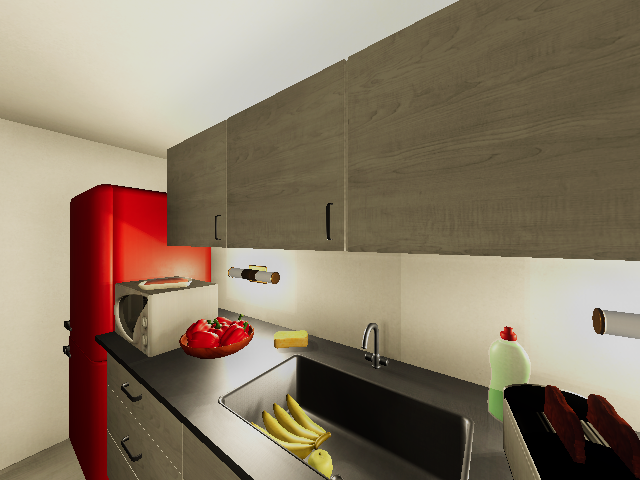}}  & \multirow{5}{*}{\begin{tabular}{c} {Apple on the sink} \\ {Bananas on the sink} \end{tabular}}
& \multirow{5}{*}{\begin{tabular}{c} 
{Locate the coffee table, take the apple, } \\ 
{pick up the bananas, find the sink, } \\
{put the apple in the sink, }\\
{put the bananas in the sink.} \end{tabular}}& \multirow{5}{*}{\begin{tabular}{c} {Prepare fruits} \\ {to serve to your guests.} \end{tabular}} \\
  \\
  \\
  \\
  \\
  \\
\multirow{5}{*}{3} & \multirow{5}{*}{\includegraphics[scale=0.86]{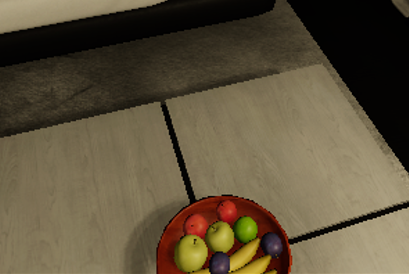}}  & \multirow{5}{*}{\begin{tabular}{c} {Apple held in hand} \\ {Bananas held in hand} \\ {Sit on sofa} \end{tabular}}
& \multirow{5}{*}{\begin{tabular}{c} {Retrieve the apple from the coffee table, } \\
{grab the bananas, } \\
{move to the sofa, sit down.}  \end{tabular}}& \multirow{5}{*}{\begin{tabular}{c} {Enjoy a fruit while}\\ {sitting on the couch.} \end{tabular}} \\
  \\
  \\
  \\
  \\
  \\
  \\ 
\multirow{5}{*}{4} & \multirow{5}{*}{\includegraphics[scale=0.18]{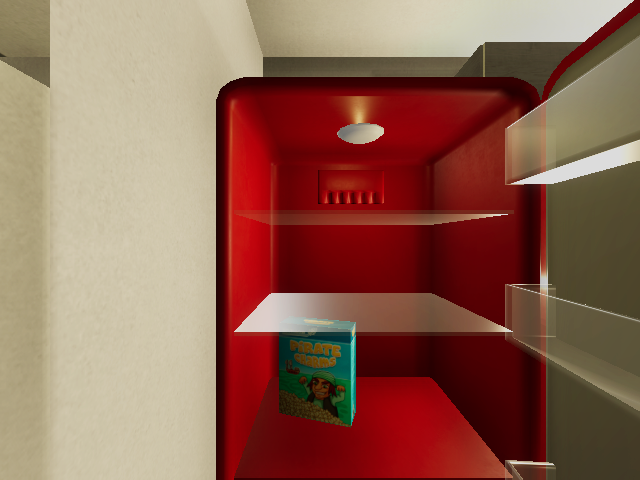}}  & \multirow{5}{*}{\begin{tabular}{c} {Cereal in fridge} \\ {Closed fridge} \end{tabular}}
& \multirow{5}{*}{\begin{tabular}{c} {Locate the cereal on the wall shelf, } \\ 
{grab it, head to the fridge,} \\
{open it, place it inside, and shut the door.}  \end{tabular}}& \multirow{5}{*}{\begin{tabular}{c} {Once breakfast is complete, } \\ 
{stow the leftovers.} \end{tabular}} \\
  \\
  \\
  \\
  \\
  \\
  \\
\multirow{5}{*}{5} & \multirow{5}{*}{\includegraphics[scale=0.18]{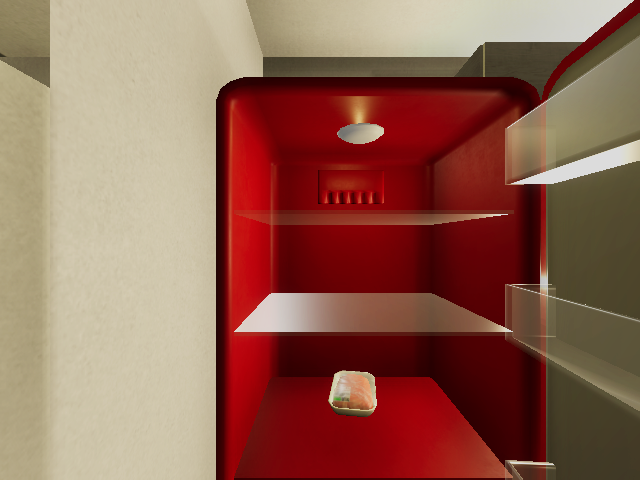}}  & \multirow{5}{*}{\begin{tabular}{c} {Salmon in the fridge} \\ {Closed fridge} \end{tabular}}
& \multirow{5}{*}{\begin{tabular}{c} {Find the salmon in the microwave, } \\
{take it to the fridge, } \\
{open the fridge, }\\
{place the salmon inside, }\\
{close the fridge.}  \end{tabular}}& \multirow{5}{*}{\begin{tabular}{c} {Store your salmon in the refrigerator} \\ {to maintain its quality.} \end{tabular}} \\
  \\
  \\
  \\
  \\
  \\
  \\ 
\multirow{5}{*}{6} & \multirow{5}{*}{\includegraphics[scale=0.18]{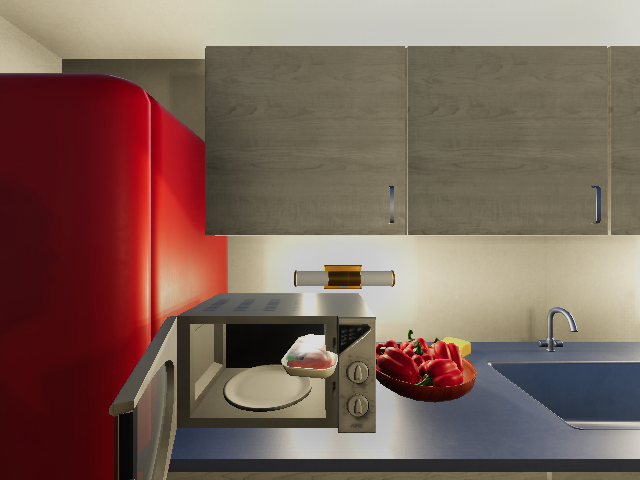}}  & \multirow{5}{*}{\begin{tabular}{c} {Salmon in the microwave} \\ {Closed microwave} \\ {Switch on the microwave} \end{tabular}}
& \multirow{5}{*}{\begin{tabular}{c} {Find the microwave, pick up the } \\ {salmon, open it, place the salmon in, } \\ {close the microwave, turn it on.} \end{tabular}}& \multirow{5}{*}{\begin{tabular}{c} {Warm up with a freshly} \\ {cooked salmon dish} \end{tabular}} \\
  \\
  \\
  \\
  \\
  \\
  \\ 
\multirow{5}{*}{7} & \multirow{5}{*}{\includegraphics[scale=0.18]{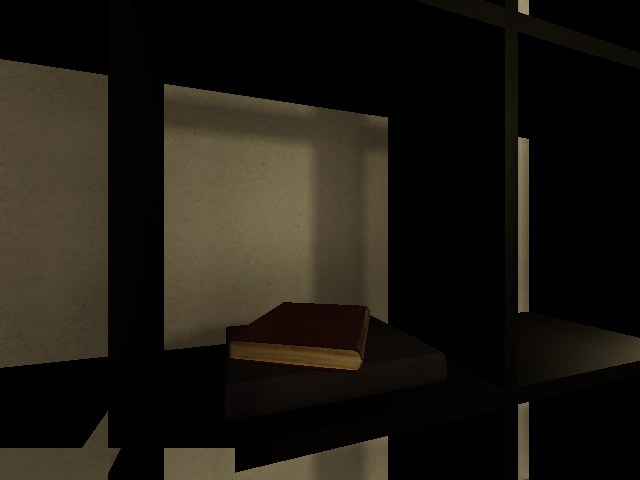}}  & \multirow{5}{*}{\begin{tabular}{c} {Book held in hand} \\ {Sit on the sofa} \end{tabular}}
& \multirow{5}{*}{\begin{tabular}{c} {Get the book from the bookshelf, } \\ 
{find the sofa, } \\
{sit on the sofa}  \end{tabular}}& \multirow{5}{*}{\begin{tabular}{c} {Take your book to the sofa} \\ {and start reading.} \end{tabular}} \\
  \\
  \\
  \\
  \\
  \\ 
\multirow{6}{*}{8} & \multirow{6}{*}{\includegraphics[scale=0.18]{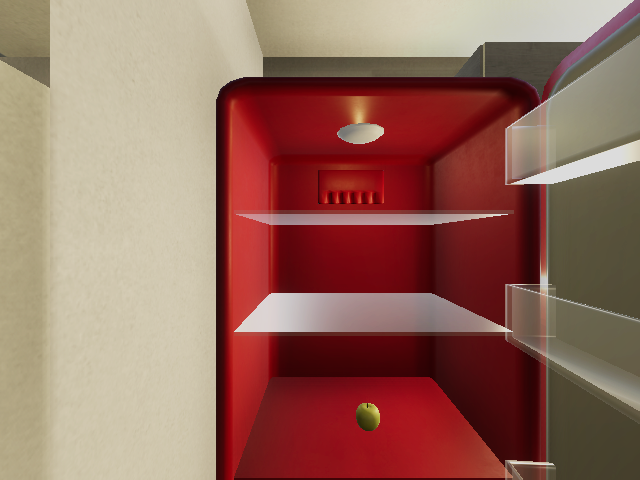}}  & \multirow{6}{*}{\begin{tabular}{c} {Apple in the fridge} \\ {Closed fridge} \end{tabular}}
& \multirow{6}{*}{\begin{tabular}{c} {find the coffee table, } \\ 
{pick up the apple, } \\
{locate the fridge, } \\
{open the fridge, } \\
{place the apple inside, } \\
{close the fridge.}  \end{tabular}}& \multirow{6}{*}{\begin{tabular}{c} {Store your apple in the fridge} \\ {for maximum freshness.}\end{tabular}} \\
  \\
  \\
  \\
  \\
  \\
  \\ 
\multirow{6}{*}{9} & \multirow{6}{*}{\includegraphics[scale=0.18]{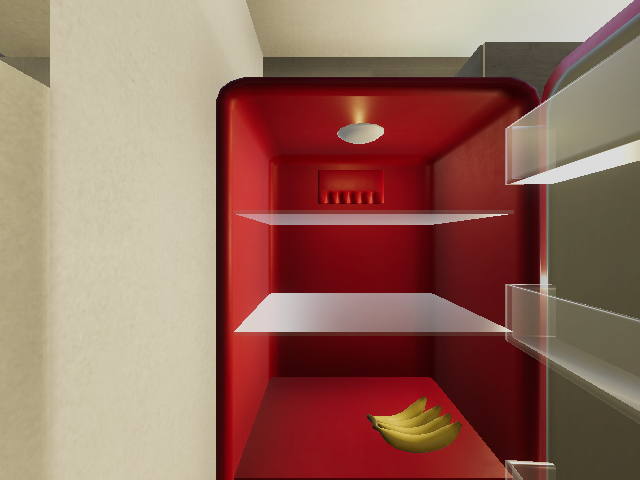}}  & \multirow{6}{*}{\begin{tabular}{c} {Bananas in the fridge} \\ {Closed fridge} \end{tabular}}
& \multirow{6}{*}{\begin{tabular}{c} {Find the coffee table,} \\ 
{grab the bananas,} \\
{locate the fridge,} \\
{open the fridge, } \\
{place the bananas inside, } \\
{close the fridge.}  \end{tabular}}& \multirow{6}{*}{\begin{tabular}{c} {Keep your bananas cool} \\ {to maintain their texture.} \end{tabular}} \\
  \\
  \\
  \\
  \\
  \\
  \\ 
\multirow{5}{*}{10} & \multirow{5}{*}{\includegraphics[scale=0.18]{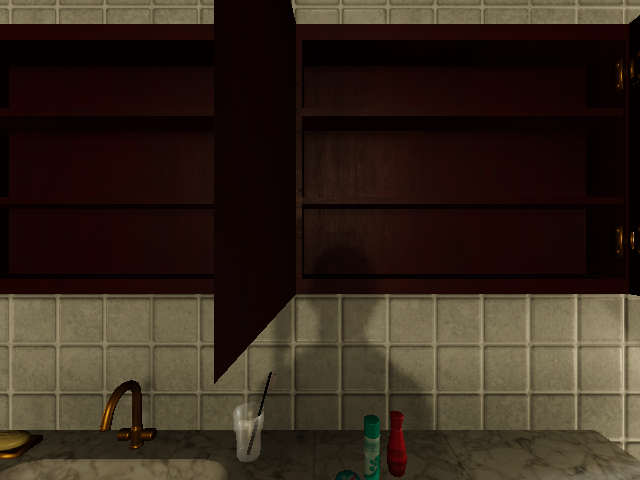}}  & \multirow{5}{*}{\begin{tabular}{c} {Toothpaste in the bathroom cabinet} \\ {Closed bathroom cabinet} \end{tabular}}
& \multirow{5}{*}{\begin{tabular}{c} {Pick up the toothpaste} \\ {from the bathroom counter,}  \\ 
{place it inside,} \\
{close the bathroom cabinet} \end{tabular}}& \multirow{5}{*}{\begin{tabular}{c} {Organize your bathroom items neatly.} \end{tabular}} \\
  \\
  \\
  \\
  \\
  \\
  \\
%   \multirow{5}{*}{11} & \multirow{5}{*}{\includegraphics[scale=0.11]{appendix_figures/table7/single/task11.png}}  & \multirow{5}{*}{\begin{tabular}{c} {clothes pants on the bathroom counter} \\ {Closed fridge}\end{tabular}}
% & \multirow{5}{*}{\begin{tabular}{c} {Find the clothes pants in the closet drawer, grab them,} \\{head to the bed, place the clothes pants on the bed.} \end{tabular}}& \multirow{5}{*}{\begin{tabular}{c}{Remove stains from your clothes } \\ {for a perfect look.} \end{tabular}} \\
%   \\
%   \\
%   \\
%   \\
%   \\
% \multirow{5}{*}{12} & \multirow{5}{*}{\includegraphics[scale=0.11]{appendix_figures/table7/single/task12.png}}  & \multirow{5}{*}{\begin{tabular}{c} {Cat on the bathtub} \end{tabular}}
% & \multirow{5}{*}{\begin{tabular}{c} {Pick up the cat from the kitchen table, go to the bathtub,} \\{ place the cat in the bathtub}  \end{tabular}}& \multirow{5}{*}{\begin{tabular}{c} {Time for a perfect clean-up for your cat.} \end{tabular}} \\
%   \\
%   \\
%   \\
%   \\
%   \\
%   \\
% \multirow{5}{*}{13} & \multirow{5}{*}{\includegraphics[scale=0.11]{appendix_figures/table7/single/task13.png}}  & \multirow{5}{*}{\begin{tabular}{c} {Cat on the bed} \\ {Sit on the bed} \end{tabular}}
% & \multirow{5}{*}{\begin{tabular}{c} {Pick up the cat from the kitchen table, go to the bathtub,} \\{ place the cat in the bathtub}  \end{tabular}}& \multirow{5}{*}{\begin{tabular}{c} {Settle into bed for a quiet evening } \\ {with your pet cat} \end{tabular}} \\
%   \\
%   \\
%   \\
%   \\
%   \\
%   \\

 \bottomrule
\end{tabular}
}
% \end{Large}
\end{center}
\caption{VirtualHome single-domain task examples}
\label{table: vh_singledomain}
\end{table*}

%% cross domain taks examples
\begin{table*}[p]
\begin{center}
\begin{Large}
\adjustbox{max width=\textwidth}{
\begin{tabular}{lc c c c cc}
\toprule
 \multirow{2}{*}{ID} & \multirow{2}{*}{Visual Observation} & \multirow{2}{*}{Goal}  & \multicolumn{2}{c}{Instruction Information} \\
 \cmidrule(rl){4-5}
 & & & Fine-grained & Abstract \\
\midrule

\multirow{5}{*}{1} & \multirow{5}{*}{\includegraphics[scale=0.18]{appendix_figures/table7/single/task4.png}}  & \multirow{5}{*}{\begin{tabular}{c} {Cereal in the fridge} \\ {Closed fridge} \end{tabular}}
& \multirow{5}{*}{\begin{tabular}{c} {Find the cereal on table} \\ {take it to the fridge,} \\ {open the door,} \\ {put it inside,} \\ {and close the refrigerator.}\end{tabular}}& \multirow{5}{*}{\begin{tabular}{c} {Once breakfast is complete,} \\ {stow the leftovers in the fridge.} \end{tabular}} \\
  \\
  \\
  \\
  \\
  \\
\multirow{5}{*}{2} & \multirow{5}{*}{\includegraphics[scale=0.18]{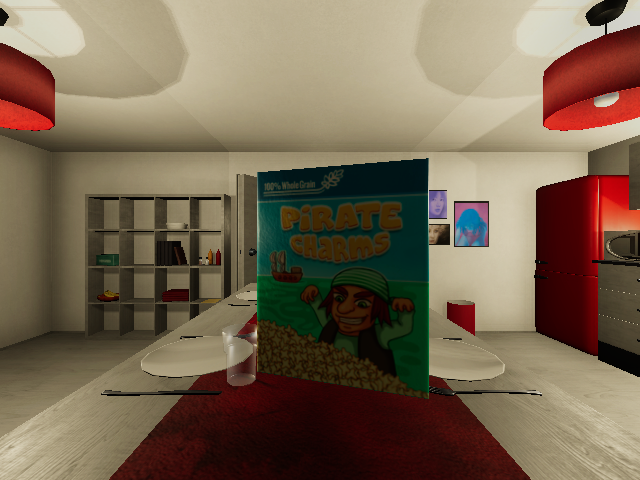}}  & \multirow{5}{*}{\begin{tabular}{c}{Cereal on kitchen table} \end{tabular}}
& \multirow{5}{*}{\begin{tabular}{c} {Pick up the cereal} \\ {from the coffee table,} \\ {move to the} \\ {kitchen table,}  \\ {set it on the table.} \end{tabular}}& \multirow{5}{*}{\begin{tabular}{c} {Place your breakfast} \\ {ready on the table.} \end{tabular}} \\
  \\
  \\
  \\
  \\
  \\
\multirow{5}{*}{3} & \multirow{5}{*}{\includegraphics[scale=0.18]{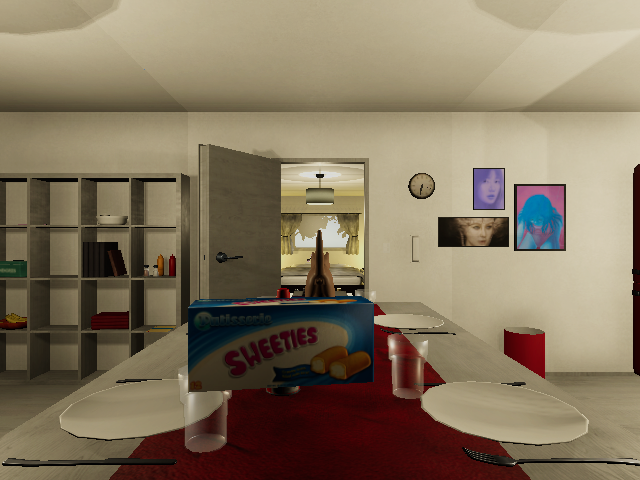}}  & \multirow{5}{*}{\begin{tabular}{c} {Creamy buns on kitchen table} \end{tabular}}
& \multirow{5}{*}{\begin{tabular}{c} {Find the coffee table,} \\ {take the creamy buns,} \\ {locate the kitchen table,} \\ {place the creamy buns} \\ {on the kitchen table.}  \end{tabular}}& \multirow{5}{*}{\begin{tabular}{c} {Organize a quick, delicious} \\ {and creamy snack on the table.} \end{tabular}} \\
  \\
  \\
  \\
  \\
  \\
  \\ 
\multirow{5}{*}{4} & \multirow{5}{*}{\includegraphics[scale=0.18]{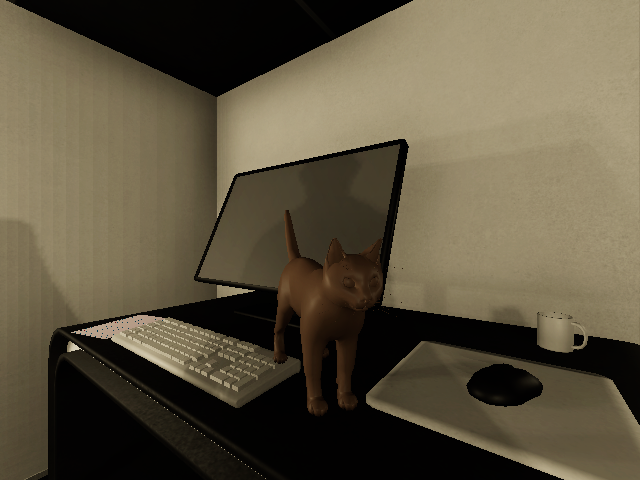}}  & \multirow{5}{*}{\begin{tabular}{c} {Cat on the desk}\end{tabular}}
& \multirow{5}{*}{\begin{tabular}{c} 
{Find the bed,} \\ {grab the cat,} \\ {locate the desk,} \\ {place the cat on the desk.}  \end{tabular}}& \multirow{5}{*}{\begin{tabular}{c} {Make your cat a part of} \\ {your workday routine.} \end{tabular}} \\
  \\
  \\
  \\
  \\
  \\
  \\
\multirow{5}{*}{5} & \multirow{5}{*}{\includegraphics[scale=0.18]{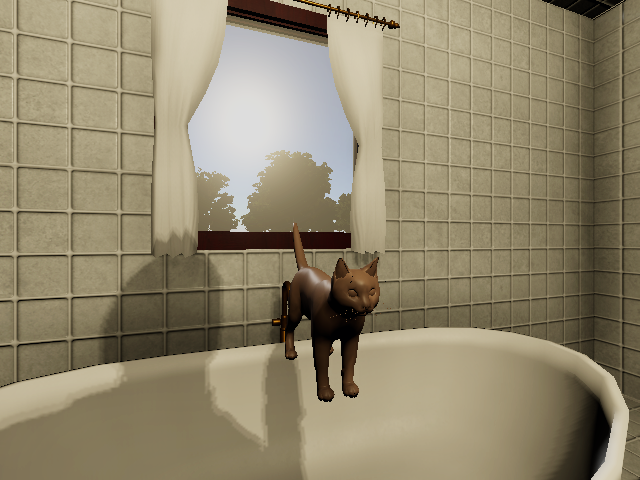}}  & \multirow{5}{*}{\begin{tabular}{c} {Cat on the bathtub}\end{tabular}}
& \multirow{5}{*}{\begin{tabular}{c}{Take the cat from the bed,} \\ {head to the bathtub,} \\{place the cat in the bathtub.}  \end{tabular}}& \multirow{5}{*}{\begin{tabular}{c} {Time for a perfect} \\ {clean-up for your cat.} \end{tabular}} \\
  \\
  \\
  \\
  \\
  \\
  \\ 
\multirow{5}{*}{6} & \multirow{5}{*}{\includegraphics[scale=0.18]{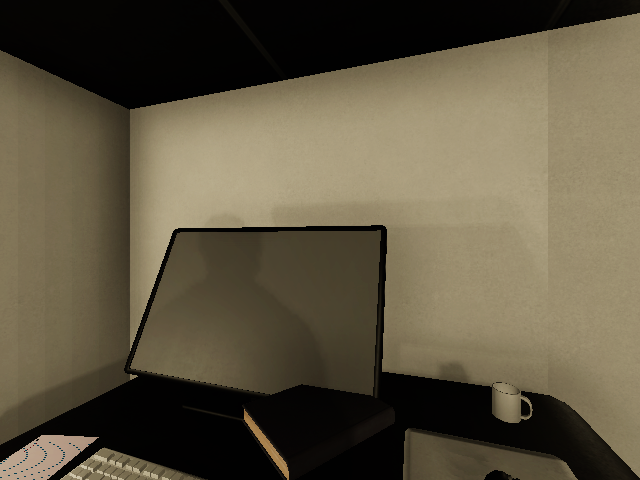}}  & \multirow{5}{*}{\begin{tabular}{c} {Book held in hand} \\ {Sit on the sofa}\end{tabular}}
& \multirow{5}{*}{\begin{tabular}{c} {Grab the book from the desk,} \\ {head to the sofa,} \\{sit down on the sofa.}  \end{tabular}}& \multirow{5}{*}{\begin{tabular}{c} {Take your book to the sofa} \\ {and start reading.} \end{tabular}} \\
  \\
  \\
  \\
  \\
  \\
  \\ 
  \multirow{5}{*}{7} & \multirow{5}{*}{\includegraphics[scale=0.18]{appendix_figures/table7/cross/task6.png}}  & \multirow{5}{*}{\begin{tabular}{c} {Book held in hand} \\ {Sit on bed} \end{tabular}}
& \multirow{5}{*}{\begin{tabular}{c} {Find the desk,} \\ {pick up the book,} \\{locate the bed,} \\ {sit down on the bed.}  \end{tabular}} & \multirow{5}{*}{\begin{tabular}{c} {Unwind before bedtime} \\ {with a soothing reading session.} \end{tabular}} \\
  \\
  \\
  \\
  \\
  \\ 
\multirow{5}{*}{8} & \multirow{5}{*}{\includegraphics[scale=0.18]{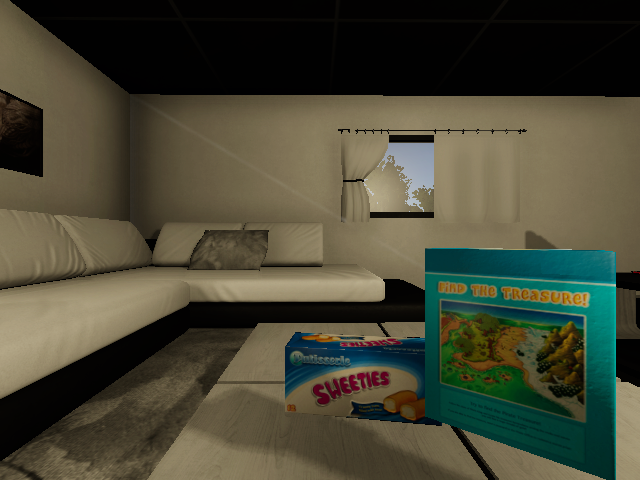}}  & \multirow{5}{*}{\begin{tabular}{c} {Creamy buns held in hand} \\ {Sit on sofa} \end{tabular}}
& \multirow{5}{*}{\begin{tabular}{c} {Find the coffee table,} \\ {grab the creamy buns,} \\{locate the sofa,} \\ {sit on the sofa.}  \end{tabular}}& \multirow{5}{*}{\begin{tabular}{c} {Indulge in a creamy bun} \\ {for a delightful sofa snack.} \end{tabular}} \\
  \\
  \\
  \\
  \\
  \\
  \\ 

 \bottomrule
\end{tabular}
}
\end{Large}
\end{center}
\vspace{-10pt}
\caption{VirtualHome cross-domain task examples}
\label{table: vh_crossdomain}
\end{table*}

\begin{table*}[h]
\centering
\resizebox{0.95\textwidth}{!}{
\begin{tabular}{>{\centering\arraybackslash}m{3cm} >{\centering\arraybackslash}m{3cm} >{\centering\arraybackslash}m{3cm} >{\centering\arraybackslash}m{3cm} >{\centering\arraybackslash}m{3cm}}
\toprule
& Cereal & Creamy Buns & Cat & Book \\ \midrule
Single-domain & \includegraphics[scale=0.11]{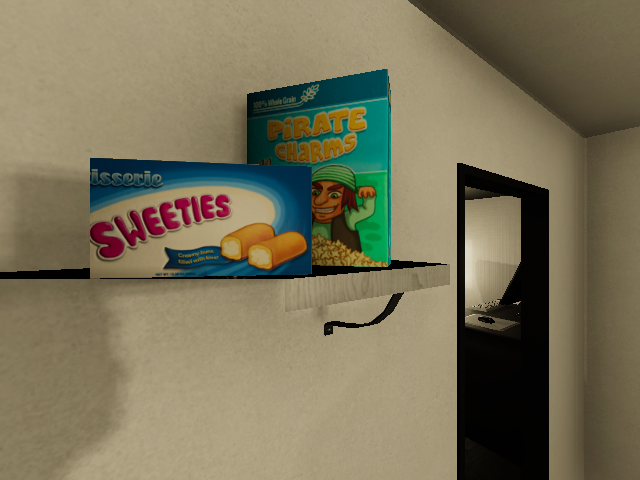} & \includegraphics[scale=0.11]{appendix_figures/table7/single/single1.png} & \includegraphics[scale=0.11]{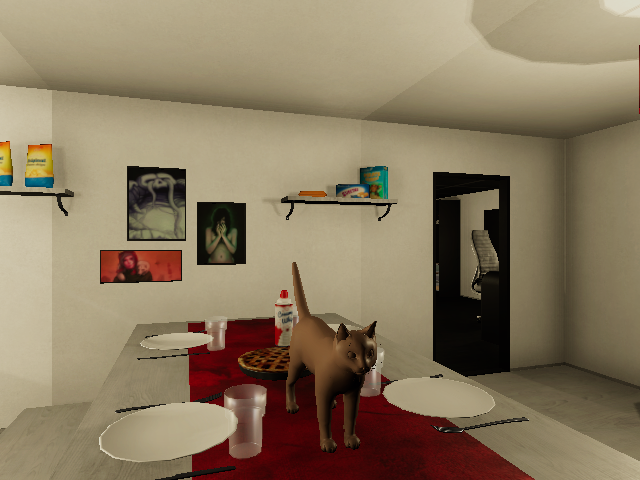} & \includegraphics[scale=0.11]{appendix_figures/table7/single/book.png} \\ 
Cross-domain & \includegraphics[scale=0.11]{appendix_figures/table7/cross/cross1.png} & \includegraphics[scale=0.11]{appendix_figures/table7/cross/cross1.png} & \includegraphics[scale=0.11]{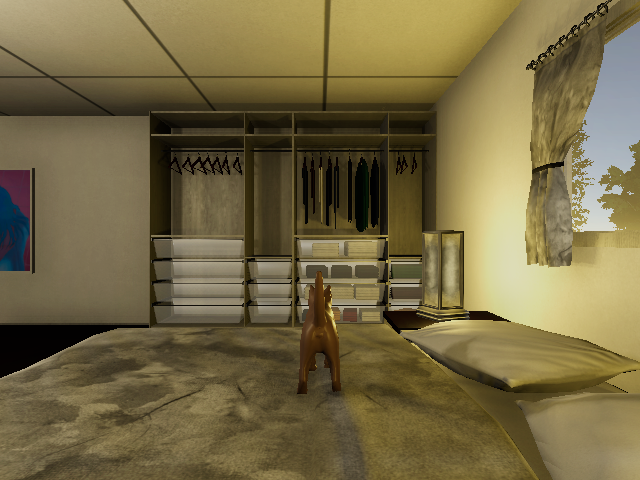} & \includegraphics[scale=0.11]{appendix_figures/table7/cross/task6.png} \\ \bottomrule
\end{tabular}
}
\vspace{-7pt}
\caption{Different object locations in cross-domain evaluation}
\label{table: cross-location}
\end{table*}

%%% LLM prompt example %%% [우리 실험에서 어떤 Prompt들을 썼는지]
\onecolumn % naive prompt
\begin{table}[h]
    \centering
    \footnotesize
    \begin{tabularx}{\textwidth}{>{\raggedright\arraybackslash}X}
        \toprule
        \textbf{Naive Prompt} \\
        \midrule
        \textbf{Robot:} Hello, I'm a robot working in a house. You can ask me to do various tasks, and I'll tell you how much each action will help you accomplish the task. I can also help you find objects relevant to the task.
        \\~\\
        \textbf{These are my scoring guidelines:}\\
        \textbf{2 points:} Actions that should follow the given previous completed actions.\\
        \textbf{1 point:} Actions that can indirectly perform or support the action that would receive 2 points.\\
        \textbf{0 points:} Actions involving visible objects that do not affect the task.\\
        \textbf{-1 point:} Actions that involve searching for objects not related to the task.\\
        \textbf{-2 points:} Actions that involve picking up or placing invisible objects, i.e., actions that cannot be performed in their current state.
        \\~\\
        Actions such as grab, put, open, sit, switch on, and close cannot be performed on invisible objects. In addition, the Close action cannot be performed if there has been no Open action in previously completed actions.
        \\~\\
        \textbf{Task Description:} State what you're trying to accomplish.\\
        \textbf{Action List:} Provide a list of the actions available in your house.\\
        \textbf{Previously Completed Actions:} List the actions that have been used previously.\\
        \textbf{Visible objects:} The objects that are currently visible to the eyes.\\
        \textbf{Grabbed:} The objects currently held in the hand.
        \\~\\
        Now you can ask for scores for actions related to the task and identify objects relevant to the task among those currently visible. I will respond in the format of the score/relevant object. Do not include any other answers, just output scores and relevant objects.
        \\~\\
        \textbf{Answer format:}\\
        Score: 2\\
        relevant objects: apple, bananas
        \\~\\
        \textbf{Human:} \\
        Task description: [Instruction].\\
        Action List: [Actions]\\
        Previously Completed Actions: [Completed Actions].\\
        Visible Objects: [Items]\\
        Grabbed: [Grabbed items]\\
        How many points is [Action]?\\
        And which of the currently visible objects is relevant to the task?
        \\~\\
        \textbf{Robot:}\\
        \bottomrule
    \end{tabularx}
    \caption{Naive prompt for LLM reward estimator $\lmgen$}
    \label{table: naiveprompt}
\end{table}

\onecolumn % CoT prompt
\begin{table}[h!]
    \centering
    \footnotesize
    \begin{tabularx}{\textwidth}{>{\raggedright\arraybackslash}X}
        \toprule
        \textbf{CoT Prompt} \\
        \midrule
        \textbf{Robot:} Hi there, I'm a robot operating in a house. You can ask me to do various tasks and I'll tell you how much each action helps in accomplishing the task. I can also help you find objects relevant to the task.
        \\~\\
        \textbf{These are my scoring guidelines:}\\
        \textbf{2 points:} Actions that should follow the given previous completed actions.\\
        \textbf{1 point:} Actions that can indirectly perform or support the action that would receive 2 points.\\
        \textbf{0 points:} Actions involving visible objects that do not affect the task.\\
        \textbf{-1 point:} Actions that involve searching for objects not related to the task.\\
        \textbf{-2 points:} Actions that involve picking up or placing invisible objects, i.e., actions that cannot be performed in their current state.
        \\~\\
        Actions such as grab, put, open, sit, switch on, and close cannot be performed on invisible objects. In addition, the Close action cannot be performed if there has been no Open action in previously completed actions.
        \\~\\
        \textbf{Task Description:} State what you're trying to accomplish.\\
        \textbf{Action List:} Provide a list of the actions available in your house.\\
        \textbf{Previously Completed Actions:} List the actions that have been used previously.\\
        \textbf{Visible objects:} The objects that are currently visible to the eyes.\\
        \textbf{Grabbed:} The objects currently held in the hand.
        \\~\\
        Now you can ask for scores for actions related to the task and identify objects relevant to the task among those currently visible.
        \\~\\
        \textbf{[Sample 1]}\\
        \textbf{Human:} \\
        Task Description: find wall shelf then grab cereal then find fridge then open fridge then put cereal in fridge then close fridge\\
        Previously Completed Actions: 1. find wall shelf\\
        Visible objects: paper, cereal, wall shelf, mouse, mug, creamy buns, crackers\\
        Grabbed: nothing\\
        \textbf{Robot:} \\
        A. Actions related to the task: [grab cereal, find fridge, open fridge, put cereal in fridge, close fridge]\\
        B. Actions that have no effect on the task (grasping a visible object): [grab creamy buns]\\
        C. Actions not related to the task: [remaining find actions]\\
        D. Interfering actions (when an item needs to be inserted but is closed or a switch is activated without closing): [none]\\
        E. Actions that cannot be performed because they are not in the visible object or are not grabbed in the completed action: [remaining actions]\\
        2 points: Among A, the action that is not in the completed action but follows it and aims to achieve the task is [grab cereal].\\
        1 point: Actions that bring results similar to those that received 2 points are [none].\\
        0 point: The actions that satisfy B are [grab creamy buns].\\
        -1 point: The actions that satisfy C are [find bookshelf, find bathtub, find sofa, find bathroom counter, find bed, find desk, find fridge, find closet drawer, find sink, find toaster, find microwave, find kitchen table, find wall shelf, find coffee table].\\
        -2 points: Actions in D, E and remaining actions.\\
        Relevant objects: wall shelf, cereal, fridge
        \\~\\
        \textbf{[Other samples]}\\
        \\~\\
        \textbf{Human:} \\
        Task description: [Instruction].\\
        Action List: [Actions]\\
        Previously Completed Actions: [Completed Actions].\\
        Visible Objects: [Items]\\
        Grabbed: [Grabbed items]\\
        How many points is [Action]?\\  % 얜 한번에 물어보는 걸로 
        \\~\\
        \textbf{Robot:}\\
        \bottomrule
    \end{tabularx}
    \caption{CoT prompt for LLM reward estimator $\lmgen$}
    \label{table: cotprompt}
\end{table}

\onecolumn % ICL prompt
\begin{table}[h!]
    \centering
    \footnotesize 
    \begin{tabularx}{\textwidth}{>{\raggedright\arraybackslash}X}
        \toprule
        \textbf{In-Context Prompt} \\
        \midrule
        \textbf{Robot:} Hi there, I'm a robot operating in a house. You can ask me to do various tasks and I'll tell you how much each action helps in accomplishing the task. I can also help you find objects relevant to the task.
        \\~\\
        \textbf{These are my scoring guidelines:}\\
        \textbf{2 points:} Actions that should follow the given previous completed actions.\\
        \textbf{1 point:} Actions that can indirectly perform or support the action that would receive 2 points.\\
        \textbf{0 points:} Actions involving visible objects that do not affect the task.\\
        \textbf{-1 point:} Actions that involve searching for objects not related to the task.\\
        \textbf{-2 points:} Actions that involve picking up or placing invisible objects, i.e., actions that cannot be performed in their current state.
        \\~\\
        Actions such as grab, put, open, sit, switch on, and close cannot be performed on invisible objects. In addition, the Close action cannot be performed if there has been no Open action in previously completed actions.
        \\~\\
        \textbf{Task Description:} State what you're trying to accomplish.\\
        \textbf{Action List:} Provide a list of the actions available in your house.\\
        \textbf{Previously Completed Actions:} List the actions that have been used previously.\\
        \textbf{Visible objects:} The objects that are currently visible to the eyes.\\
        \textbf{Grabbed:} The objects currently held in the hand.
        \\~\\
        Now you can ask for scores for actions related to the task and identify objects relevant to the task among those currently visible.
        \\~\\
        \textbf{[Sample 1]}\\ 
        \textbf{Human:} \\
        \textbf{Task Description:} find wall shelf then grab cereal then find fridge then open fridge then put cereal in fridge then close fridge\\
        \textbf{Previously Completed Actions:} 1. find wall shelf\\
        \textbf{Visible objects:} paper, cereal, wall shelf, mouse, mug, creamy buns, crackers\\
        \textbf{Grabbed:} nothing\\
        \textbf{Robot:} \\
        grab cereal: 2
        \\~\\
        \textbf{[Other samples]}\\
        \\~\\
        \textbf{Human:} \\
        \textbf{Task Description:} <Instruction>\\
        \textbf{Action List:} <Actions>\\
        \textbf{Previously Completed Actions:} <Completed actions>\\
        \textbf{Visible Objects:} <Visible Objects>\\
        \textbf{Grabbed:} <Grabbed Objects>\\
        How many points is <Action>?\\
        \\~\\
        \textbf{Robot:} \\
        \bottomrule
    \end{tabularx}
    \caption{ICL prompt for LLM reward estimator $\lmgen$}
    \label{table: iclprompt}
\end{table}

\onecolumn % ICL prompt2
\begin{table}[h!]
    \centering
    \footnotesize 
    \begin{tabularx}{\textwidth}{>{\raggedright\arraybackslash}X}
        \toprule
        \textbf{In-Context Prompt (2)} \\
        \midrule
        \textbf{Objective:} To successfully achieve your goal, execute a sequence of actions listed below. The order of execution should be logical and based on the situation provided. Only use actions from the specified action set for decision-making and scoring. Any actions not listed are not to be considered for this task.
        \\~\\
        \textbf{Scoring Guidelines:}\\
        \textbf{2 Points (Highly Beneficial Action):} Awarded to a single action that is crucial for directly achieving the goal, delivering immediate and substantial benefits, and can be executed in its current state.\\
        \textbf{1 Point (Beneficial Action):} Allocated to actions that are significant steps or preparatory actions toward the goal, facilitating notable progression or preparation, and are executable in their current state.\\
        \textbf{0 Points (Neutral Action):} Given to actions that are either indirectly related to the goal or have minimal contribution towards its achievement, essentially actions that are tangential to the current objective, but still executable in their current state.\\
        \textbf{-1 Point (Potentially Detrimental Action):} Assigned to actions that, without directly blocking the goal, can indirectly impede its achievement, squander time on activities unrelated to the objective, or cannot be executed in their current state.\\
        \textbf{-2 Points (Directly Detrimental Action):} Awarded to actions that directly interfere with goal achievement or have an effect opposite to the intended goal.
        \\~\\
        \textbf{Task Description:} Specify the goal you're trying to achieve.\\
        \textbf{Action List:} Action list\\
        \textbf{Previously Completed Actions:} List the actions that have been used previously.\\
        \textbf{Visible objects:} The objects currently visible to the eyes. Find the objects relevant to the task description among these objects.\\
        \textbf{Grabbed:} The objects currently being held in the hand.
        \\~\\
        \textbf{[Sample 1]}\\ 
        \textbf{Task Description:} find wall shelf then grab cereal then find fridge then open fridge then put cereal in fridge then close fridge\\
        \textbf{Previously Completed Actions:} 1. find wall shelf\\
        \textbf{Visible objects:} paper, cereal, wall shelf, mouse, mug, creamy buns, crackers\\
        \textbf{Grabbed:} nothing\\
        \textbf{Response:} grab cereal: 2
        \\~\\
        \textbf{[Other samples]}\\
        \\~\\
        \textbf{Human:} \\
        \textbf{Task Description:} <Instruction>\\
        \textbf{Action List:} <Actions>\\
        \textbf{Previously Completed Actions:} <Completed actions>\\
        \textbf{Visible Objects:} <Visible Objects>\\
        \textbf{Grabbed:} <Grabbed Objects>\\
        \textbf{Response:} \\
        \bottomrule
    \end{tabularx}
    \caption{ICL prompt (2) for LLM reward estimator $\lmgen$}
    \label{table: iclprompt2}
\end{table}

\begin{table*}[t]
    \centering
    \begin{adjustbox}{width=0.8 \textwidth}
    \begin{tabular}{c >{\centering\arraybackslash}p{3cm} >{\centering\arraybackslash}p{3cm} >{\centering\arraybackslash}p{3cm}}
    \specialrule{.1em}{.05em}{.05em}
    \textbf{Instruction} & \multicolumn{3}{c}{Enjoy a fruit snack while sitting on the couch.} \\
    \textbf{Observation} & \multicolumn{3}{c}{picture frame} \\
    \textbf{Action} & \multicolumn{3}{c}{grab apple} \\    \textbf{Execution History} & \multicolumn{3}{c}{None} \\
    \textbf{Rewards} & $r^C=-2$ & $r^S=2$ & $r^T=-2$ \\   

    \cmidrule(lr){1-1}\cmidrule(lr){2-4}

    Prompt $\mathcal{P}_1$ & 2 \checkmark & 2 \ding{55} & 2 \checkmark \\
    Prompt $\mathcal{P}_2$ & 1 \checkmark & 1 \ding{55} & 1 \ding{55} \\
    Prompt $\mathcal{P}_3$ & -2 \checkmark & -2 \ding{55} & -2 \checkmark \\
    Prompt $\mathcal{P}_4$ & -2 \checkmark & -2 \checkmark & -2  \checkmark\\
    Prompt $\mathcal{P}_5$ & 2 \checkmark & 2 \ding{55} & 2  \checkmark\\

    \specialrule{.1em}{.05em}{.05em}
    \end{tabular}
    \end{adjustbox}
    \caption{An example of how reward estimation differs according to contextual, structural, and temporal consistency. In each consistency-based reward ($r^C$, $r^S$, and $r^T$), a check mark indicates that the predicted reward contributes to the majority voting for its respective consistency. An 'X' mark signifies that the reward is disregarded due to either failing the backward-verification process (in temporal consistency) or incorrectly responding to the MDP-specific query (in structural consistency).}
    \label{table: consistency_rewards1}
\end{table*}

\begin{table*}[t]
    \centering
    \begin{adjustbox}{width=0.8 \textwidth}
    \begin{tabular}{c >{\centering\arraybackslash}p{3cm} >{\centering\arraybackslash}p{3cm} >{\centering\arraybackslash}p{3cm}}
    \specialrule{.1em}{.05em}{.05em}
    \textbf{Instruction} & \multicolumn{3}{c}{Enjoy the crisp, refreshing taste of a wholesome apple sandwich.} \\
    \textbf{Observation} & \multicolumn{3}{c}{dish washing liquid, bread slice, coffee pot, stove, bell pepper, sink, fridge} \\
    \textbf{Action} & \multicolumn{3}{c}{put apple on bread slice} \\
    \textbf{Execution History} & \multicolumn{3}{c}{1. find coffee table 2. grab apple, 3. find toaster, 4. grab bread slice} \\
    \textbf{Rewards} & $r^C=-2$ & $r^S=-2$ & $r^T=2$ \\   

    \cmidrule(lr){1-1}\cmidrule(lr){2-4}

    Prompt $\mathcal{P}_1$ & 2 \checkmark & 2  \ding{55} & 2 \checkmark \\
    Prompt $\mathcal{P}_2$ & -2 \checkmark & -2   \checkmark & -2 \checkmark  \\
    Prompt $\mathcal{P}_3$ & 2 \checkmark & 2  \checkmark & 2 \ding{55} \\
    Prompt $\mathcal{P}_4$ & -2 \checkmark & -2 \checkmark & -2  \checkmark\\
    Prompt $\mathcal{P}_5$ & 1 \checkmark & 1  \checkmark & 1  \checkmark\\

    \specialrule{.1em}{.05em}{.05em}
    \end{tabular}
    \end{adjustbox}
    \caption{An example of how reward estimation differs according to contextual, structural, and temporal consistency.}
    \label{table: consistency_rewards2}
\end{table*}

\begin{table*}[t]
    \centering
    \begin{adjustbox}{width=0.8 \textwidth}
    \begin{tabular}{c >{\centering\arraybackslash}p{3cm} >{\centering\arraybackslash}p{3cm} >{\centering\arraybackslash}p{3cm}}
    \specialrule{.1em}{.05em}{.05em}
    \textbf{Instruction} & \multicolumn{3}{c}{Prepare for bath time with your cat.} \\
    \textbf{Observation} & \multicolumn{3}{c}{cat, bathtub, tower} \\
    \textbf{Action} & \multicolumn{3}{c}{find bathtub} \\
    \textbf{Execution History} & \multicolumn{3}{c}{1. find kitchen table 2. grab cat, 3. find bathtub, 4. put cat in bathtub} \\
    \textbf{Rewards} & $r^C=-1$ & $r^S=2$ & $r^T=-1$ \\   

    \cmidrule(lr){1-1}\cmidrule(lr){2-4}

    Prompt $\mathcal{P}_1$ & -1 \checkmark &  -1  \checkmark& -1 \checkmark \\
    Prompt $\mathcal{P}_2$ & 2 \checkmark &  2   \checkmark & 2 \checkmark  \\
    Prompt $\mathcal{P}_3$ & -1 \checkmark & -1  \checkmark & -1 \checkmark\\
    Prompt $\mathcal{P}_4$ & 2 \checkmark & 2 \checkmark & 2  \checkmark\\
    Prompt $\mathcal{P}_5$ & 2 \checkmark & 2  \ding{55} & 2  \checkmark\\

    \specialrule{.1em}{.05em}{.05em}
    \end{tabular}
    \end{adjustbox}
    \caption{An example of how reward estimation differs according to contextual, structural, and temporal consistency.}
    \label{table: consistency_rewards3}
\end{table*}

\end{document}